\documentclass[journal]{IEEEtran}

\usepackage{graphicx} 
\usepackage{makecell}
\usepackage{subfigure}
\usepackage{color}
\usepackage{bm}
\usepackage{setspace}
\usepackage[ruled]{algorithm2e}
\SetKwRepeat{Do}{do}{while}
\usepackage{multirow,subfigure,pifont,hyperref}
\usepackage{verbatim}
\usepackage{amsthm,amsmath,amssymb}
\usepackage{mathrsfs}
\RequirePackage{amsmath}[2016/11/05]

\begin{document}
	\title{How Image Generation Helps Visible-to-Infrared Person Re-Identification?}
	\author{Honghu Pan, Yongyong Chen, Member,~IEEE, Yunqi He, Xin Li*, Zhenyu He*,~\IEEEmembership{Senior Member,~IEEE} 
		\thanks{This research is supported in part by the National Natural Science
			Foundation of China (Grant No.62172126 and Grant No.62106063), by the Shenzhen Research Council (Grant No. JCYJ20210324120202006), by the Guangdong Natural Science Foundation under Grant 2022A1515010819, by the Shenzhen College Stability Support Plan (Grant GXWD20201230155427003-20200824113231001), and by The Major Key Project of PCL (Grant PCL2021A03-1).}
		\thanks{H. Pan is with School of Computer Science and Technology, Harbin Institute of Technology, Shenzhen, Shenzhen 518055, China. (Email: 19B951002@stu.hit.edu.cn)}
		\thanks{Y. Chen is with School of Computer Science and Technology, Harbin Institute of Technology, Shenzhen 518055, China, and also with Guangdong Provincial Key Laboratory of Novel Security Intelligence Technologies. (Email: YongyongChen.cn@gmail.com)}
		\thanks{Y. He is with College of Information and Computer Engineering, Northeast Forestry University, Harbin 150000, China. (Email: heyunqi.cs@gmail.com)}
		\thanks{X. Li is with Peng Cheng Laboratory, Shenzhen 518055, China. (Email: xinlihitsz@gmail.com)}
		\thanks{Z. He is with School of Computer Science and Technology, Harbin Institute of Technology, Shenzhen, Shenzhen 518055, China, and also with Peng Cheng Laboratory, Shenzhen 518055, China. (Email: zhenyuhe@hit.edu.cn)}
	}
	
	\maketitle
	
	\begin{abstract}
		Compared to visible-to-visible (V2V) person re-identification (ReID), the visible-to-infrared (V2I) person ReID task is more challenging due to the lack of sufficient training samples and the large cross-modality discrepancy.
		To this end, we propose Flow2Flow, a unified framework that could jointly achieve training sample expansion and cross-modality image generation for V2I person ReID.
		Specifically, Flow2Flow learns bijective transformations from both the visible image domain and the infrared domain to a shared isotropic Gaussian domain with an invertible visible flow-based generator and an infrared one, respectively.
		With Flow2Flow, we are able to generate pseudo training samples by the transformation from latent Gaussian noises to visible or infrared images, and generate cross-modality images by transformations from existing-modality images to latent Gaussian noises to missing-modality images.
		For the purpose of identity alignment and modality alignment of generated images, we develop adversarial training strategies to train Flow2Flow.
		Specifically, we design an image encoder and a modality discriminator for each modality.
		The image encoder encourages the generated images to be similar to real images of the same identity via identity adversarial training, and the modality discriminator makes the generated images modal-indistinguishable from real images via modality adversarial training.
		Experimental results on SYSU-MM01 and RegDB demonstrate that both training sample expansion and cross-modality image generation can significantly improve V2I ReID accuracy.
	\end{abstract}


	\section{Introduction}
	Person re-identification (ReID), which aims to match pedestrian images captured by non-overlapped cameras, is a crucial technique in video surveillance.
	In recent years, the person ReID methods~\cite{TriNet,BoT} have achieved human-level accuracy on some large-scale datasets~\cite{market1501,MSMT17}.
	However, these methods assume that the pedestrian images are captured by visible-spectrum cameras under bright environments, and do not work well in the nighttime surveillance scenarios.
	Considering that the infrared radiation is immune to illumination, the visible-to-infrared (V2I) person ReID~\cite{SYSU,RegDB,MPANet,cmGAN}, which denotes a cross-spectrum or cross-modality matching task, has gained a broad attention in the computer vision community.

	\begin{figure}[t]
		\centering
		\begin{center}
			\includegraphics[width=0.45 \textwidth]{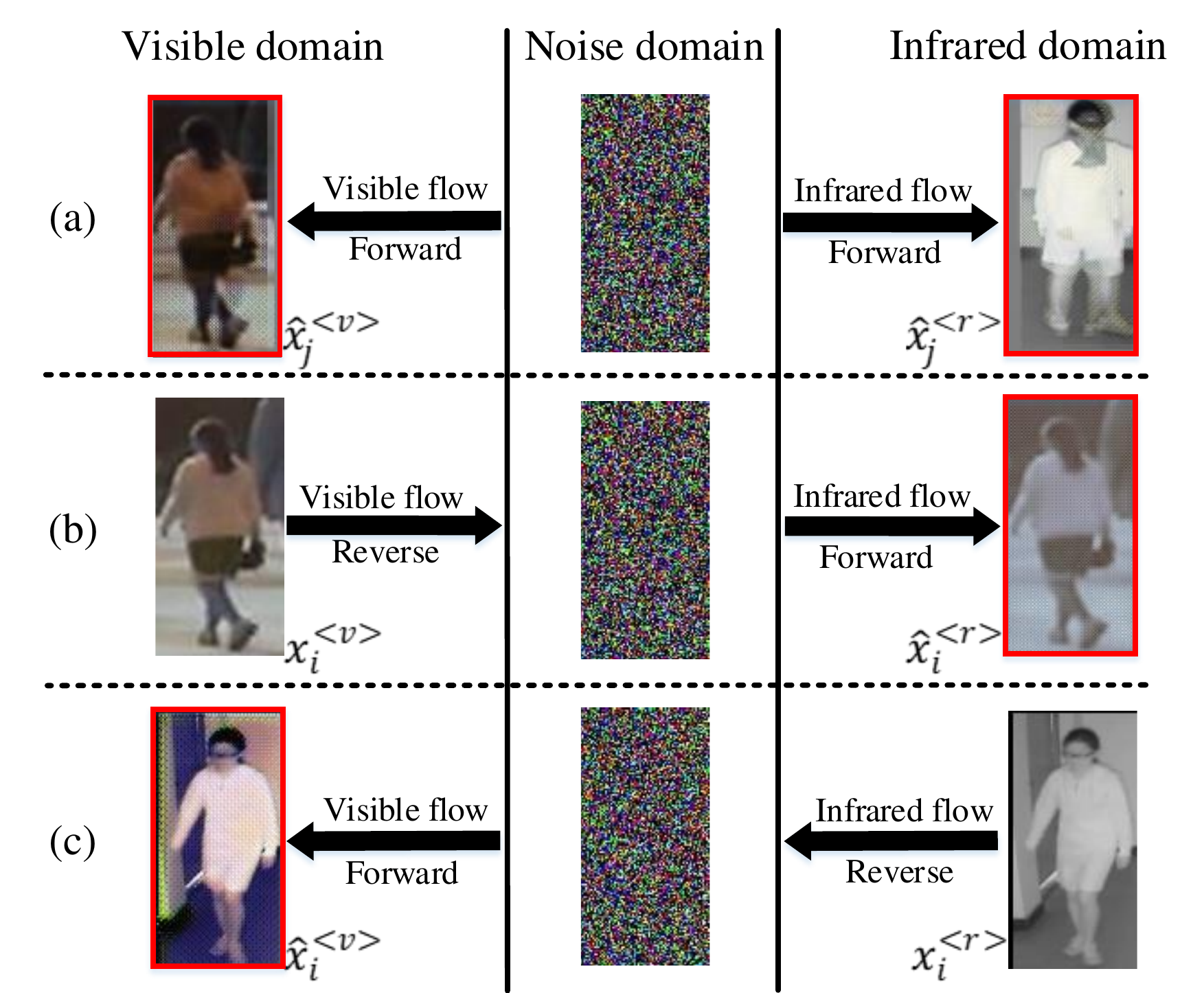}
		\end{center}
		\caption{Schematic of (a) training sample generation, (b) visible-to-infrared cross-modality image generation, and (c) infrared-to-visible cross-modality image generation, in which images outlined by red boxes are generated images.
			The proposed Flow2Flow contains a visible flow and an infrared flow, which learn bijective transformations from the visible image domain and infrared domain to an isotropic Gaussian domain, respectively.
			(a) Training sample generation: a latent Gaussian noise could be transformed to pseudo visible sample $\hat{x}^{<v>}_j$ or pseudo infrared sample $\hat{x}^{<r>}_j$ by the forward propagation of visible flow or infrared flow.
			(b) Cross-modality generation from visible domain to infrared domain: the visible image $x^{<v>}_i$ is first transformed to a latent Gaussian noise $z_i$ by reverse propagation of the visible flow, then $z_i$ can be transformed to the corresponding infrared image $\hat{x}^{<r>}_i$ by forward propagation of the infrared flow.
			(c) Vice versa for cross-modality generation from infrared domain to visible domain.
		}
		\label{fig_generation}
	\end{figure}

	Although recent researches~\cite{MPANet,cmGAN,FMCNet} have made great efforts on V2I ReID, it is still very challenging due to the following two reasons.
	First, the number of training images in V2I datasets~\cite{SYSU, RegDB} is not as large as that in visible-to-visible (V2V) ReID datasets~\cite{market1501,MSMT17}, especially for the infrared images.
	For example, MSMT17~\cite{MSMT17}, one of large-scale V2V datasets, contains 32,621 training samples, while SYSU-MM01~\cite{SYSU} and RegDB~\cite{RegDB} only contain 9,929 and 2,060 infrared images for training, respectively.
	Second, the modality discrepancy between the visible spectrum and infrared spectrum is very large.
	In V2V ReID, some of recent studies~\cite{JVTC,PN-GAN} have considered to generate pseudo training samples to reduce intra-class variance, while to our knowledge, training sample expansion has not been studied in V2I ReID.
	Existing methods of V2I ReID~\cite{AlignGAN,cmGAN,cmPIG} mainly struggle to reduce the cross-modality discrepancy via cross-modality image generation, most of which employ the generative adversarial network (GAN)~\cite{GAN,CycleGAN} to generate the missing modality images for existing modality images.

	In this paper, we develop Flow2Flow, a unified framework to explore how image generation, including the training sample generation and cross-modality image generation, improves the V2I person ReID task.
	Specifically, our framework contains two flow-based generative models~\cite{RealNVP,Glow}, i.e., a visible flow and an infrared flow, which learn invertible or bijective transformations from the visible image domain and infrared image domain to an isotropic Gaussian domain, respectively.
	Thereby, generating pseudo visible or infrared training samples could be achieved by the forward flow propagation from the latent noise domain to the visible or infrared image domain.
	While generating missing-modality images from given-modality images could be achieved by transformations from given-modality domain to Gaussian noise domain to missing-modality domain.
	Fig.~\ref{fig_generation} shows the schematic of the training sample generation and cross-modality image generation.
	
	To guarantee the invertibility and exact log-likelihood computation, existing flow models~\cite{RealNVP,Glow} is composed of mutiple $1 \times 1$ convolutional layers and linear coupling layers, which leads to insufficient nonlinearity. 
	To resolve this, we implement an extra invertible activation layer in the last block of the visible and infrared flows to increase model nonlinearity.
	In addition, we propose an identity adversarial training strategy and a modality adversarial training strategy to encourage the generated images corresponding to specific identities and modalities.
	For the purpose of adversarial training, we implement two discriminators for each modality, including an image encoder for identity alignment and a modality discriminator for modality alignment.
	To enable the identity alignment of the real images and generated images, we minimize the distance between their encoded features when training generators, and maximize that when training discriminators.
	While the modality discriminators distinguish whether the images are generated or from a specific real modality.
	
	To the best of our knowledge, this is the first study that achieves both training sample generation and cross-modality generation via a unified framework.
	Experimental results demonstrate that both generations improve the V2I ReID performance significantly.
	For example, the training sample expansion and cross-modality generation obtain gains of 2.0\% and 1.2\% mAP against the baseline model on the SYSU-MM01~\cite{SYSU} dataset.
	The main contributions of this paper are three-fold:
	\begin{itemize}
		\item To explore how image generation helps V2I person ReID, we propose Flow2Flow, a unified framework, to jointly generate pseudo training samples and cross-modality images, which contains a visible flow and an infrared flow to learn bijective transformations from image domains to Gaussian domain;
		\item For the purpose of identity alignment and modality alignment of generated images, we develop an image encoder and a modality discriminator for each modality to perform the identity adversarial training and modality adversarial training, respectively;
		\item We demonstrate that both the training sample expansion and cross-modality generation improve the V2I ReID accuracy significantly. In addition, our Flow2Flow model leads a new state-of-the-arts (SOTA) performance on the SYSU-MM01 dataset.
	\end{itemize}
	
	The remainder of this paper is organized as follows:
	Section~\ref{related_Works} introduces recent literatures related to this paper;
	Section~\ref{preliminary} simply reviews theoretical backgrounds of the flow-based generative models;
	Section~\ref{method} elaborates the Flow2Flow model in detail; 
	Section~\ref{experiments} presents the ablation studies, visualizations and comparisons with the SOTA;
	Section~\ref{conclusion} draws brief conclusions.

	
	\section{Related Works}
	\label{related_Works}

	\subsection{Visible-to-Visible Person ReID}
The V2V person ReID is a single-modality image retrieval task, which devotes to enlarging the inter-class variance and reducing the intra-class variance.
To this end, existing methods mainly consider three levels of factors: objective-level, network-level and data-level.
For the objectives or loss functions, TriNet~\cite{TriNet} proposed the hard triplet mining strategy on the basis of triplet loss to learn pedestrian representations; 
BoT~\cite{BoT} combined the cross entropy loss and triplet loss to train network;
moreover, the center loss~\cite{centerlearning} and angular loss~\cite{VA-reID} have also been successfully applied in the V2V person ReID.
For the network, early works~\cite{TriNet} learned the global features from pedestrian images via a single CNN branch.
Next, the multi-branch architecture has been adopted to learn the multi-granularity or part-level features~\cite{MGN,PCB,AAGCN}.
Furthermore, data augmentation or generation~\cite{JTVC,PN-GAN} could also improve the ReID accuracy, which belongs to the data-based category.
For example, PN-GAN~\cite{PN-GAN} generated multi-pose pedestrian images via GAN model, which could reduce the pedestrian view variance;
JVTC~\cite{JTVC} conducted the online data augmentation for contrastive learning, in which the mesh projections were taken as the references to generate multi-view images.

	\subsection{Visible-to-Infrared Person ReID}
	\label{V2IPR}
	The V2I person ReID enables the cross-spectrum pedestrian retrieval, whose crux is to reduce the large cross-modality discrepancy.
	Existing V2I ReID methods 
	mainly have two techniques to reduce the modal discrepancy: 1) learning the modality-shared pedestrian representation and 2) compensating information of missing modality via generative models~\cite{GAN,CycleGAN}.
	The modality-shared ones~\cite{DGD,expAT,eBDTR,MPANet} projected the visible and infrared pedestrian images into a shared Euclidean space, in which the intra-class similarity and inter-class similarity are maximized and minimized, respectively.
	For example, DGD-MSR~\cite{DGD} proposed a modality-specific network to extract	modality-specific representations from each modality; expAT~\cite{expAT} devised an exponential angular triplet loss beyond the Euclidean metric based constraints to learn the angularly discriminative features;
	MPANet~\cite{MPANet} aimed to capture the nuances of cross-modality images via a modality alleviation module and a pattern alignment module.
	The modality compensation ones~\cite{FMCNet,DDRL,AlignGAN,cmGAN,cmPIG} usually generated missing modality information from existing modality data:
	DDRL~\cite{DDRL} proposed an image-level sub-network based on GAN model, which could translate a visible (infrared) image to a corresponding infrared (visible) one;
	cmPIG~\cite{cmPIG} employed the set-level alignment information to generate instance alignment cross-modality paired-images;
	FMCNet~\cite{FMCNet} utilized the feature-level modality compensation to reduce modality discrepancy, which generated the cross-modality features rather than images.
	The method proposed in this paper could be classified as the modality compensation category.
	Compared to existing methods that directly learn a transformation from given modality to missing modality via GAN models, our method employs the flow-based generative models to construct invertible transformations from given modality to latent Gaussian noise to missing modality.
	Thereby, besides the cross-modality generation, our method could generate pseudo training samples via transformations from Gaussian noise to image modalities.

	\subsection{Flow-based Generative Model}
	The flow-based generative model constructs an invertible or bijective mapping from the complex distribution of true data to a simple distribution (e.g., isotropic Gaussian distribution).
	For the purpose of invertibility and exact log-likelihood computation, layers in flow-based model should be carefully designed.
	RealNVP~\cite{RealNVP} proposed the affine coupling layer, which could easily compute the determinant of Jaocibian matrix;
	Glow~\cite{Glow} presented an invertible $1\times 1$ convolution layer, meanwhile the LU decomposition was utilized to speed up the computation of determinants;
	cAttnFlow~\cite{cAttnFlow} introduced the invertible attentions to increase the nonlinearity of flow-based model.
	Recently, a great number of works have extended the flow-based model into speech synthesis~\cite{Waveglow}, molecular graph generation~\cite{GraphAF,MoFlow} and image generation~\cite{Glow,Srflow,Hcflow}.
	For the molecular graph generation, MoFlow~\cite{MoFlow} implemented an atom flow and a conditional bond flow to generate the atom features and atom bonds in molecular, respectively.
	For the image super-resolution, SRFlow~\cite{Srflow} and HCFlow~\cite{Hcflow} took the low-resolution images as the condition, and thus learned the high-resolution images via a conditional flow.
	In this paper, we take advantage of the invertibility of flow-based model to achieve 1) generating pseudo samples from isotropic Gaussian noises and 2) cross-modality image generation from existing modality to latent noises to missing modality.
	As far as we can tell, this is the first study that applies the flow-based model in person ReID.
	
	\subsection{Generative Adversarial Network}
	The first GAN model was proposed in ~\cite{GAN}, which consists of a generator and a discriminator, and they could improve each other by the adversarial training.
	In GAN model, the generator generates samples from noise variables with a known probability density function (PDF) and tries to fool the discriminator, and the discriminator distinguishes whether the data is true or fake to beat the generator.
	Recently, the GAN architectures have been heavily refined to adapt various application scenarios.
	For instance, the Conditional GAN~\cite{CGAN,SGAN} could generate samples corresponding to specific condition labels; CycleGAN~\cite{CycleGAN} enabled the unpaired cross-domain image translation by the cycle consistency loss.
	Meanwhile, the GAN model also showed its priority in the V2I person ReID~\cite{FMCNet,DDRL,AlignGAN} and V2I person ReID areas~\cite{JTVC,PN-GAN}.
	Unlike the flow-based model~\cite{RealNVP,Glow} which could exactly compute the log-likelihood of true data, GAN model implicitly minimizes the KL divergence between the true data and data generated from noises.
	To make the generated data indistinguishable from the real data, training a GAN model pursues an equilibrium between the generator and discriminator, which requires careful experimental setup tuning.
	In this paper, we combine the flow-based model and adversarial training to generate the high-quality visible and infrared pedestrian images.

	\begin{figure*}[t]
		\centering
		\begin{center}
			\includegraphics[width=0.97 \textwidth]{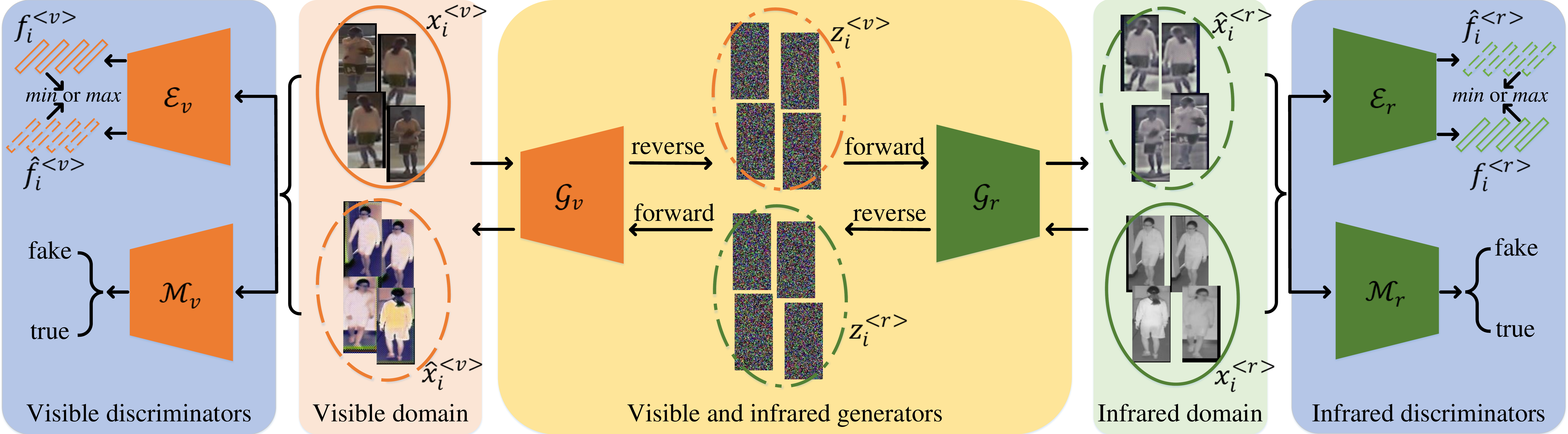}
		\end{center}
		\caption{Framework of Flow2Flow. It consists of visible generator $\mathcal{G}_v$, infrared generator $\mathcal{G}_r$, visible encoder $\mathcal{E}_v$, infrared encoder $\mathcal{E}_r$, visible modality discriminator $\mathcal{M}_v$ and infrared modality discriminator $\mathcal{M}_r$:
			$\mathcal{G}_v$ or $\mathcal{G}_r$ learns a bijective transformation from the visible domain $P(X^{<v>})$ or infrared domain $P(X^{<r>})$ to the latent Gaussian domain $\Pi(Z)$;
			$\mathcal{E}_v$ and $\mathcal{E}_r$ encourage the generated images to be similar to true images of the same identity via identity adversarial training; 
			$\mathcal{M}_v$ and $\mathcal{M}_r$ make the generated images modal-indistinguishable from real images via modality adversarial training.
			To generate pseudo training images of specific identity, we first perform the linear interpolation of latent noises with the desired identity, then generate pseudo training samples via forward propagation from noises to images.
			To achieve the cross-modality generation, we exploit the invertibility of flow-based model, i.e., $P(X^{<v>}) \rightarrow \Pi(Z) \rightarrow P(X^{<r>})$ and $P(X^{<r>}) \rightarrow \Pi(Z) \rightarrow P(X^{<v>})$.
		}
		\label{fig_framework}
	\end{figure*}
	
	\section{Preliminaries}
	\label{preliminary}
	The flow-based generative model aims to learn a bijective transformation from a complex distribution $X \sim P(X)$ to a simple distribution $Z \sim \Pi(Z)$ with a known probability density function, in which $X$ denotes the true training data and $\Pi(Z)$ is usually a Gaussian distribution.
	For the purpose of bijective mapping, the flow-based model consists of a sequence of invertible generators $\mathcal{G}=\mathcal{G}_1 \bullet \cdots \bullet \mathcal{G}_L$:
	\begin{equation}
	x_i = \mathcal{G}(z_i), z_i = \mathcal{G}^{-1}(x_i).
	\label{eq_x_and_z}
	\end{equation}
	By the change of variable formula, $P(X)$ and $\Pi(Z)$ satisfy the following transformation:
	\begin{equation}
	P(X) = \Pi(Z) \left| \det (J_{\mathcal{G}^{-1}}) \right|,
	\label{eq_Px_Pz}
	\end{equation}
	where $\mathrm{det} (J_{\mathcal{G}^{-1}})$ denotes the determinant of Jacobian matrix.
	Then the objective of $\max \{\log (P(X))\}$ can be converted to:
	\begin{equation}
	\max \{ \sum_{i} \log (\Pi(z_i)) + \sum_{l=1}^L \log \left| \det (J_{\mathcal{G}^{-1}_l}) \right| \}.
	\label{eq_flow_objective}
	\end{equation}
	From Eq.~(\ref{eq_x_and_z}), Eq.~(\ref{eq_Px_Pz}) and Eq.~(\ref{eq_flow_objective}), we could know that the training process of the flow-based model follows the reverse propagation, and the inference or generation process follows the forward propagation.
	
	A standard flow-based model mainly contains two categories of layers: invertible $1\times 1$ convolution layer~\cite{Glow} and affine coupling layer~\cite{Nice,RealNVP}.
	For a single generator $\mathcal{G}_l$ in $\mathcal{G}$, the reverse and forward projection of the $1\times 1$ convolution layer has the following expression:
	\begin{equation}
	z_i^{<l-1>} = W_l z_i^{<l>}, z_i^{<l>} = W_l^{-1} z_i^{<l-1>},
	\end{equation}
	where $Z^{<0>}$ and $Z^{<L>}$ denotes $Z$ and $X$, respectively.
	The design of the affine coupling layer should allow 1) invertible transformation and 2) exact computation of the Jacobian determinant $\det (J_{\mathcal{G}^{-1}})$. Its reverse and forward transformation from $\mathbb{R}^n$ to $\mathbb{R}^n$ can be denoted as:
	\begin{equation}
	\begin{split}
	&z_{i(1:d)}^{<l-1>} = z_{i(1:d)}^{<l>}, \\
	z_{i(d+1:n)}^{<l-1>} = z_{i(1:d)}^{<l>} \odot &{\rm sigmoid}(S_{\Theta}(z_{i(1:d)}^{<l>})) + T_{\Theta}(z_{i(1:d)}^{<l>});
	\end{split}
	\label{eq_affine_coupling_forward}
	\end{equation}
	\begin{equation}
	\begin{split}
	&z_{i(1:d)}^{<l>} = z_{i(1:d)}^{<l-1>} , \\
	z_{i(d+1:n)}^{<l>} = (z_{i(d+1:n)}^{<l-1>}-&T_{\Theta}(z_{i(1:d)}^{<l-1>})) / {\rm sigmoid}(S_{\Theta}(z_{i(1:d)}^{<l-1>})).
	\end{split}
	\label{eq_affine_coupling_back}
	\end{equation}
	In Eq.(~\ref{eq_affine_coupling_forward}) and Eq.(~\ref{eq_affine_coupling_back}), $S_{\Theta}$ and $T_{\Theta}$ are learning-based neural networks, and $\rm{sigmoid}$ is the activation function.
	Readers could refer to the origin papers~\cite{Glow,Nice,RealNVP} for more algorithm details.
	
	\begin{table}[t]
		\small
		\centering
		\caption{Notations and descriptions.}
		\begin{spacing}{1.25}
			\begin{tabular}{cl}
				\hline
				Notations & Descriptions \\ \hline
				$x^{<v>}_i$, $x^{<r>}_i$	& The true visible and infrared images         \\  
				$\hat{x}^{<v>}_i$, $\hat{x}^{<r>}_i$ & The generated visible and infrared images          \\
				$y^{<v>}_i$, $y^{<r>}_i$	& The identity labels of $x^{<v>}_i$ and $x^{<r>}_i$         \\  
				$\mathcal{G}_v$, $\mathcal{G}_r$ & The visible and infrared generators            \\
				$z^{<v>}_i$, $z^{<r>}_i$ & The latent noises learned from $x^{<v>}_i$ and $x^{<r>}_i$            \\
				$\mathcal{E}_v$, $\mathcal{E}_r$ & The visible and infrared image encoders            \\
				$f^{<v>}_i$, $f^{<r>}_i$ & The encoded features of $x^{<v>}_i$ and $x^{<r>}_i$            \\
				$\hat{f}^{<v>}_i$, $\hat{f}^{<r>}_i$ & The encoded features of $\hat{x}^{<v>}_i$ and $\hat{x}^{<r>}_i$            \\
				$\mathcal{M}_v$, $\mathcal{M}_r$ & The visible and infrared modality discriminators            \\
				\hline
			\end{tabular}	
		\end{spacing}
		\label{table_notation}
	\end{table}

	\section{Methodologies}
	\label{method}
	In this section, we first introduce the architecture of the proposed Flow2Flow model in Section~\ref{MA}, then present the training objective (including training generators and discriminators) in Section~\ref{OfMT}, and finally elaborate the training sample generation and cross-modality generation procedures in Section~\ref{IG}.

	\subsection{Model Architecture}
	\label{MA}
	This paper aims to combine the flow-based generative model and adversarial training to jointly achieve expansion of training samples and cross-modality image generation.
	To this end, we devise a Flow2Flow model as shown in Fig.\ref{fig_framework}, which consists of a visible flow $\mathcal{G}_v$, an infrared flow $\mathcal{G}_r$, visible encoder $\mathcal{E}_v$, infrared encoder $\mathcal{E}_r$, visible modality discriminator $\mathcal{M}_v$ and infrared modality discriminator $\mathcal{M}_r$.
	To make this paper clear, we present the notations and their corresponding descriptions in Table~\ref{table_notation}.

	\begin{table*}[t]
		\centering
		\small
		\caption{The reverse propagation, forward propagation and log-determinant of three main components in the nonlinear flow.}
		\begin{tabular}{cccc}
			\hline
			Description                 & Reverse propagation & Forward propagation & log-determinant \\ \hline
			Invertible convolution layer~\cite{Glow}      & $z_i^{<l-1>} = W_l z_i^{<l>}$                    & $z_i^{<l>} = W_l^{-1} z_i^{<l-1>}$                    & $ \sum(\log \left| \det(W_l) \right| )$              \\ \hline
			Affine coupling layer~\cite{RealNVP}       & \makecell{$z_{i(1:d)}^{<l-1>} = z_{i(1:d)}^{<l>}$  \\ $s={\rm sigmoid}(S_{\Theta}(z_{i(1:d)}^{<l>}))$ \\ $t=T_{\Theta}(z_{i(1:d)}^{<l>})$ \\ $z_{i(d+1:n)}^{<l-1>} = z_{i(1:d)}^{<l>} \odot s + t$}
			& \makecell{$z_{i(1:d)}^{<l>} = z_{i(1:d)}^{<l-1>}$  \\ $s={\rm sigmoid}(S_{\Theta}(z_{i(1:d)}^{<l-1>}))$ \\ $t=T_{\Theta}(z_{i(1:d)}^{<l-1>})$ \\ $z_{i(d+1:n)}^{<l>} = (z_{i(1:d)}^{<l-1>}-t) / s $}                    & $ \sum( \log \left| s \right|) $                \\ \hline
			Nonlinear activation layer & $z_i^{<l-1>} = \frac{\exp\{z_i^{<l>}\}-\exp\{-z_i^{<l>}\}}{\exp\{z_i^{<l>}\}-\exp\{-z_i^{<l>}\}}$                    & $z_i^{<l>} = \frac{1}{2} \log \frac{1+z_i^{<l-1>}}{1-z_i^{<l-1>}} $                    & $ \sum( \log \frac{4 \cdot \exp\{2 \cdot z_i^{<l>}\}} {{(\exp\{2 \cdot z_i^{<l>}\}+1)}^2} )$                \\ \hline
		\end{tabular}
		\label{table_flow_layer}
	\end{table*}

	In our Flow2Flow architecture, the visible models ($\mathcal{G}_v$, $\mathcal{E}_v$ and $\mathcal{M}_v$) and their infrared counterparts have the same network structure but do not share weights.
	Motivated by GAN models, we adopt the adversarial training to generate high-quality fake images: $\mathcal{G}_v$ and $\mathcal{G}_r$ denote the generators, while encoders $\mathcal{E}_v$ and $\mathcal{E}_r$, modality discriminators $\mathcal{M}_v$ and $\mathcal{M}_r$ refer to as the discriminator models.
	
	For the generators $\mathcal{G}_v$ and $\mathcal{G}_r$, we first implement multiple invertible $1\times 1$ convolution layers and affine coupling layers as the standard flow-based model~\cite{RealNVP,Glow} of Section~\ref{preliminary}.
	Moreover, to increase the nonlinearity of the standard flow, we propose an extra invertible nonlinear activation layer.
	Specifically, for the reverse propagation from $Z^{<l>}$ to $Z^{<l-1>}$, the nonlinear layer follows \textit{tanh} activation function:
	\begin{equation}
	z_i^{<l-1>} = \frac{\exp\{z_i^{<l>}\}-\exp\{-z_i^{<l>}\}}{\exp\{z_i^{<l>}\}+\exp\{-z_i^{<l>}\}}.
	\label{eq_nonlinear_rev}
	\end{equation}
	The forward propagation from $Z^{<l-1>}$ to $Z^{<l>}$ has the following expression:
	\begin{equation}
	z_i^{<l>} = \frac{1}{2} \log \frac{1+z_i^{<l-1>}}{1-z_i^{<l-1>}}.
	\label{eq_nonlinear_for}
	\end{equation}
	We call $\mathcal{G}_v$ and $\mathcal{G}_r$ with the activation layer as the nonlinear flow.
	The reverse propagation, forward propagation and log-determinant of three main components in $\mathcal{G}_v$ and $\mathcal{G}_r$ are summarized in Table~\ref{table_flow_layer}.
	
	The encoders $\mathcal{E}_v$ and $\mathcal{E}_r$ aim to achieve the identity alignment for true and fake images with the same identity. They are composed of multiple CNN layers, which encode the true image $x_i$ and fake image $\hat{x}_i$ as $f_i$ and $\hat{f}_i$, respectively.
	We devise an identity adversarial training strategy to train $\mathcal{E}_v$ and $\mathcal{E}_r$: in the stage of training generators, we minimize the distance between $f_i$ and $\hat{f}_i$ if they correspond the same identity label $y_i$; while in the stage of training discriminators, we maximize the distance between them.
	The modality discriminators $\mathcal{M}_v$ and $\mathcal{M}_r$ aim to achieve the modality alignment for true and fake images.
	Compared to $\mathcal{E}_v$ and $\mathcal{E}_r$, $\mathcal{M}_v$ and $\mathcal{M}_r$ have one more CNN layer to map the features into binary modality logits.
	Motivated by CycleGAN~\cite{CycleGAN}, we devise an modality adversarial training strategy to train $\mathcal{M}_v$ and $\mathcal{M}_r$:
	when training generators, we hope that the generated visible image $\hat{x}^{<v>}_i$ could be classified to the real visible modality by $\mathcal{M}_v$;
	when training discriminators, $\mathcal{M}_v$ struggles to classify $x^{<v>}_i$ and $\hat{x}^{<v>}_i$ as the real visible modality and fake visible modality, respectively.
	And vice versa for $\mathcal{M}_r$.
	The detailed training objectives are presented in Section~\ref{OfMT}.

	\subsection{Objectives for Model Training}
	\label{OfMT}
	Our Flow2Flow architecture consists of two categories models: 1) the flow-based generators $\mathcal{G}_v$ and $\mathcal{G}_r$ that map the true images into latent noise; 2) discriminators ($\mathcal{E}_v$, $\mathcal{E}_r$, $\mathcal{M}_v$ and $\mathcal{M}_r$) that encourage generators to generate images corresponding to specific identities and specific modalities via adversarial training.
	
	\vspace{2mm}
	\textbf{Training flow-based generators.}
	By the flow objective of Eq.~(\ref{eq_flow_objective}), to maximize the log-likelihood of the training data, we need to maximize the log-likelihood of latent noises and the log-determinants of Jacobian matrices.
	Maximizing log-likelihood of latent noises is equivalent to minimizing its negative log-likelihood (NLL).
	The NLL of Gaussian distribution $\Pi(Z)$ can be denoted as $\frac{m}{2}\log(2\pi) + m \log(\sigma) + \frac{1}{2 \sigma^2} \sum{{(z_i-\mu)}^2}$, where $m$, $\mu$ and $\sigma$ are the number of samples, mean and standard deviation, respectively.
	Here we omit the first two terms since they are constant terms.
	Then the flow loss can be denoted as:
	\begin{equation}
	\begin{split}
	L_{flow} & = \frac{1}{2 \sigma^2} \sum{{(z_i^{<v>}-\mu)}^2}  - \sum \log \left| \det (J_{\mathcal{G}^{-1}_v}) \right| \\
	&  + \frac{1}{2 \sigma^2} \sum{{(z_i^{<r>}-\mu)}^2}  - \sum \log \left| \det (J_{\mathcal{G}^{-1}_r}) \right|.
	\end{split}
	\label{eq_loss_flow}
	\end{equation}
	
	In addition to maximizing $\log (P(X))$, we add a cluster constraint on the latent noises, which encourages noises with the same identity to be close to each other.
	Specifically, we minimize the distance between intra-class noises and maximize that between inter-class ones:
	\begin{equation}
	L_{noise} = \frac{1}{n_1} \sum_{y_i=y_j} d(z_i,z_j) - \frac{1}{n_2} \sum_{y_i\neq y_k} d(z_i,z_k),
	\label{eq_loss_noise}
	\end{equation}	
	where $z_i$ can be $z_i^{<v>}$ or $z_i^{<r>}$; $d(\cdot,\cdot)$ indicates the Euclidean distance; $n_1$ and $n_2$ denote the number of intra-class pairs and inter-class pairs, respectively.
	Then the total generator loss $L_{\mathcal{G}}$ for training the visible and infrared flows can be defined as the combination of the flow loss and latent noise loss:
	\begin{equation}
	L_{\mathcal{G}} = L_{flow} + \lambda L_{noise}.
	\label{eq_loss_G}
	\end{equation}
	
	\vspace{2mm}
	\textbf{Identity adversarial training of image encoders.}
	The encoders $\mathcal{E}_v$ and $\mathcal{E}_r$ belong to the discriminator models, thereby their weights are frozen when training generators.
	To achieve the set-level alignment, $\mathcal{E}_v$ and $\mathcal{E}_r$ encourage the fake images generated by $\mathcal{G}_v$ and $\mathcal{G}_r$ could be similar to true images of the same identity:
	\begin{equation}
	\begin{split}
	L_{\mathcal{E}}^{\mathcal{G}} & = \frac{1}{n_1^{<v>}} \sum_{y^{<v>}_i=y^{<r>}_j} d \left ( \mathcal{E}_v (x^{<v>}_i), \mathcal{E}_v (\mathcal{G}_v(\mathcal{G}^{-1}_r(x^{<r>}_j))) \right ) \\
	& + \frac{1}{n_1^{<r>}} \sum_{y^{<r>}_i=y^{<v>}_j} d \left (\mathcal{E}_r (x^{<r>}_i), \mathcal{E}_r (\mathcal{G}_r(\mathcal{G}^{-1}_v(x^{<v>}_j))) \right ). 
	\end{split}
	\label{eq_loss_encoder_G}
	\end{equation}
	In Eq.~(\ref{eq_loss_encoder_G}), $\mathcal{E}(\cdot)$ denotes the encoded feature, such as $f^{<v>}_i=\mathcal{E}_v(x^{<v>}_i)$ and $\hat{f}^{<v>}_j=\mathcal{E}_v(\mathcal{G}_v(\mathcal{G}^{-1}_r(x^{<r>}_j)))$; $n_1^{<v>}$ and $n_1^{<r>}$ denote the number of intra-class visible pairs and intra-class infrared pairs, respectively.
	
	In the stage of training discriminators, we freeze the weights of $\mathcal{G}_v$ and $\mathcal{G}_r$ and update the weights of $\mathcal{E}_v$ and $\mathcal{E}_r$.
	At this time, we minimize the similarity between the true images and generated images of the same identity:
	\begin{equation}
	\begin{split}
	L_{\mathcal{E}}^{\mathcal{D}} & = 2 - \frac{1}{n_1^{<v>}} \sum_{y^{<v>}_i=y^{<r>}_j} d(\mathcal{E}_v (x^{<v>}_i), \mathcal{E}_v (\hat{x}^{<v>}_j)) \\
	& + 2 - \frac{1}{n_1^{<r>}} \sum_{y^{<r>}_i=y^{<v>}_j} d(\mathcal{E}_r (x^{<r>}_i), \mathcal{E}_r (\hat{x}^{<r>}_j)) ,
	\end{split}
	\label{eq_loss_encoder_D}
	\end{equation}
	in which features learned by the encoders are normalized to unit-length, so that the distance is within the interval [0, 2].
	
	\vspace{2mm}
	\textbf{Modality adversarial training of modality discriminators.}
	When training generators, the weights of $\mathcal{M}_v$ and $\mathcal{M}_r$ are frozen.
	We hope the generated image $\hat{x}^{<v>}_i$ or $\hat{x}^{<r>}_i$ can be classified to the visible or infrared modality by $\mathcal{M}_v$ or $\mathcal{M}_r$:
	\begin{equation}
	\begin{split}
	L_{\mathcal{M}}^{\mathcal{G}} & = (1-\mathcal{M}_v(\mathcal{G}_v(\mathcal{G}^{-1}_r(x^{<r>}_i)))) \\
	& + (1-\mathcal{M}_r(\mathcal{G}_r(\mathcal{G}^{-1}_v(x^{<v>}_i)))),
	\end{split}
	\label{eq_loss_discriminator_G}
	\end{equation}
	in which $\mathcal{M}_v$ and $\mathcal{M}_r$ output the modality logits.
	While in the stage of training discriminators, the modality discriminators $\mathcal{M}_v$ and $\mathcal{M}_r$ struggle to classify the true images and generated ones as 1 and 0, respectively:
	\begin{equation}
	\begin{split}
	L_{\mathcal{M}}^{\mathcal{D}} & = \left ( (1-\mathcal{M}_v(x^{<v>}_i) + (0-\mathcal{M}_v(\hat{x}^{<v>}_i) \right ) \\
	& + \left ( (1-\mathcal{M}_r(x^{<r>}_i) + (0-\mathcal{M}_r(\hat{x}^{<r>}_i) \right ).
	\end{split}
	\label{eq_loss_discriminator_D}
	\end{equation}

	\subsection{Image Generation}
	\label{IG}	
	Thanks to the invertibility property of the flow-based models, Flow2Flow of Fig.\ref{fig_framework} could jointly achieve training sample expansion and cross-modality image generation.
	
	\vspace{2mm}
	\textbf{Training sample expansion} aims to generate images corresponding to specific identities.
	To this end, we fully exploit the invertibility of flow and latent space interpolation technique.
	For visible images $x_{i_1}^{<v>}$ and $x_{i_2}^{<v>}$ corresponding to identity label $y_i^{<v>}$, we first feed them into $\mathcal{G}_v$ for reverse propagation to learn their respective latent noise $z_{i_1}^{<v>}$ and $z_{i_2}^{<v>}$:
	\begin{equation}
	z_{i_1}^{<v>} = \mathcal{G}_v^{-1}(x_{i_1}^{<v>}), z_{i_2}^{<v>} = \mathcal{G}_v^{-1}(x_{i_2}^{<v>}).
	\label{eq_xv2zv}
	\end{equation}
	We then obtain a fake visible image $\hat{x}^{<v>}_i$ corresponding to identity $y_i^{<v>}$ by the following equation:
	\begin{equation}
	\hat{x}^{<v>}_i = \mathcal{G}_v(z_{i_1}^{<v>}+\frac{p}{q}(z_{i_2}^{<v>}-z_{i_1}^{<v>})),
	\label{eq_zv2xv_hat}
	\end{equation}
	where $p,q\in N^+$ and $q>p\geq1$.
	For simplicity, we omit the process of generating fake infrared image $\hat{x}^{<r>}_i$ from $x^{<v>}_i$.

	\vspace{2mm}
	\textbf{Cross-modality image generation} aims to generate images of missing modality from images of existing modality.
	Given a visible image $x_i^{<v>}$, we first learns its latent noise $z_i^{<v>}$ by the reverse propagation of $\mathcal{G}_v$, then generate its corresponding infrared image $\hat{x}^{<r>}_i$ by the forward propagation of $\mathcal{G}_r$:
	\begin{equation}
	z_i^{<v>}=\mathcal{G}_v^{-1}(x_i^{<v>}), \hat{x}^{<r>}_i=\mathcal{G}_r(z_i^{<v>}).
	\label{eq_xv2zv2xr}
	\end{equation}
	Similarly, we could generate a visible image $\hat{x}^{<v>}_i$ from a given infrared  image $x_i^{<r>}$ by the following equations:
	\begin{equation}
	z_i^{<r>}=\mathcal{G}_r^{-1}(x_i^{<r>}), \hat{x}^{<v>}_i=\mathcal{G}_v(z_i^{<r>}).
	\label{eq_xr2zv2xv}
	\end{equation}

	\section{Experiments}
	\label{experiments}
	
	\subsection{Experimental Settings}
	\label{DaEI}
	
	\textbf{Benchmarks.}
	To validate the effectiveness of Flow2Flow, we conduct our experiments on two widely-used V2I ReID datasets: SYSU-MM01~\cite{SYSU} and RegDB~\cite{RegDB}. 
	SYSU-MM01 is currently the largest V2I ReID dataset, which is composed of 491 identities captured by 4 visible and 2 infrared cameras.
	Its training set consists of 20,284 visible images and 9,929 infrared images from 296 persons, and the query set contains 3,803 infrared images from 96 identities.
	RegDB dataset is composed of 4,120 visible images and 4,120 infrared images from 412 identities, where each identity contains 10 visible images and 10 infrared images; it is randomly and evenly divided into the training set and the testing set.

	\textbf{Implementations.}
	In Flow2Flow, the visible flow $\mathcal{G}_v$ and infrared flow $\mathcal{G}_r$ contains 12 invertible blocks, in which each block have an affine coupling layer and a $1 \times 1 $ convolution layer.
	We add the invertible activation layer in the last block to increase the nonlinearity of $\mathcal{G}_v$ and $\mathcal{G}_r$.
	The image encoders $\mathcal{E}_v$ and $\mathcal{E}_r$ is composed of four convolutional layers that encode the input images into 512-dimensional features.
	While the modal discriminators $\mathcal{M}_v$ and $\mathcal{M}_r$ have one more convolutional layer to learn the binary modality logits.
	We train Flow2Flow 50 epochs with Adam optimizer~\cite{Adam}, whose learning rate is set to $2 \times 10^{-4}$.
	For each iteration of the training stage, we alternately train the generators twice and train discriminators once.
	Meanwhile, the input images are resized to $144 \times 72$.
	And $\lambda$ in Eq.(~\ref{eq_loss_G}) is set to 0.01.
	
	For the ReID model, we choose MPANet~\cite{MPANet}, the current SOTA model, as our baseline model.
	We adopt the same experimental settings as baseline for fair comparison.
	Different from cmPIG~\cite{cmPIG} that directly concatenates the true images and generated images, we separately learn features of true images and generated images, then perform the feature-level concatenation.
	For the evaluation metrics, we report mAP and Rank1 of CMC.

	\begin{table*}[ht]
		\small
		\centering
		\renewcommand\arraystretch{1.1}
		\caption{Comparison with SOTA on SYSU-MM01~\cite{SYSU} and RegDB~\cite{RegDB}. 
			We compare two categories of methods: (1) modality-shared methods and (2) modality compensation ones.
			The best performance is marked in \textcolor{red}{red}.
		}
		\begin{tabular}{cc|cccccccc|cccc}
			\hline
			\multicolumn{2}{c|}{\multirow{4}{*}{Method}}             & \multicolumn{8}{c|}{SYSU-MM01}                                                                                                                 & \multicolumn{4}{c}{RegDB}                                                                                      \\ \cline{3-14} 
			\multicolumn{2}{c|}{}                                    & \multicolumn{4}{c|}{All-Search}                                         & \multicolumn{4}{c|}{Indoor-Search}                                   & \multicolumn{2}{c|}{\multirow{2}{*}{Visible2Infrared}} & \multicolumn{2}{c}{\multirow{2}{*}{Infrared2Visible}} \\ \cline{3-10}
			\multicolumn{2}{c|}{}                                    & \multicolumn{2}{c|}{Single-Shot}   & \multicolumn{2}{c|}{Multi-Shot}    & \multicolumn{2}{c|}{Single-Shot}   & \multicolumn{2}{c|}{Multi-Shot} & \multicolumn{2}{c|}{}   & \multicolumn{2}{c}{}              \\ \cline{3-14} 
			\multicolumn{2}{c|}{}                                    & Rank1 & \multicolumn{1}{c|}{mAP}   & Rank1 & \multicolumn{1}{c|}{mAP}   & Rank1 & \multicolumn{1}{c|}{mAP}   & Rank1          & mAP            & Rank1                   & \multicolumn{1}{c|}{mAP}     & Rank1                             & mAP               \\ \hline
			\multicolumn{1}{c|}{\multirow{8}{*}{(1)}} & ZeroPadding~\cite{SYSU} & 14.80 & \multicolumn{1}{c|}{15.95} & 19.13 & \multicolumn{1}{c|}{10.89} & 20.58 & \multicolumn{1}{c|}{26.92} & 24.43          & 18.86          & -                       & \multicolumn{1}{c|}{-}       & -                                 & -                 \\
			\multicolumn{1}{c|}{}                     & DDAG~\cite{DDAG}        & 54.75 & \multicolumn{1}{c|}{53.02} & -     & \multicolumn{1}{c|}{-}     & 61.02 & \multicolumn{1}{c|}{67.98} & -              & -              & 69.34                   & \multicolumn{1}{c|}{63.46}   & 68.06                             & 61.80             \\
			\multicolumn{1}{c|}{}                     & expAT~\cite{expAT}        & 38.57 & \multicolumn{1}{c|}{38.61} & 44.71 & \multicolumn{1}{c|}{32.20} & -     & \multicolumn{1}{c|}{-}     & -              & -              & 67.54                   & \multicolumn{1}{c|}{66.51}   & 66.48                             & 64.31             \\
			\multicolumn{1}{c|}{}                     & NFS~\cite{NFS}          & 56.91 & \multicolumn{1}{c|}{55.45} & 63.51 & \multicolumn{1}{c|}{48.56} & 62.79 & \multicolumn{1}{c|}{69.79} & 70.03          & 61.45          & 80.54                   & \multicolumn{1}{c|}{72.10}   & 77.95                             & 69.79             \\
			\multicolumn{1}{c|}{}                     & CAJL~\cite{CAJL}         & 69.88 & \multicolumn{1}{c|}{53.61} & -     & \multicolumn{1}{c|}{-}     & 76.26 & \multicolumn{1}{c|}{76.79} & -              & -              & 85.03                   & \multicolumn{1}{c|}{65.33}   & -                                 & -                 \\
			\multicolumn{1}{c|}{}                     & SPOT~\cite{SPOT}         & 65.34 & \multicolumn{1}{c|}{62.25} & -     & \multicolumn{1}{c|}{-}     & 69.42 & \multicolumn{1}{c|}{70.48} & -              & -              & 80.35                   & \multicolumn{1}{c|}{72.46}   & 79.37                             & 72.26             \\
			\multicolumn{1}{c|}{}                     & DTRM~\cite{DTRM}         & 63.03 & \multicolumn{1}{c|}{58.63} & -     & \multicolumn{1}{c|}{-}     & 66.35 & \multicolumn{1}{c|}{71.76} & -              & -              & 79.09                   & \multicolumn{1}{c|}{70.09}   & 78.02                             & 69.56             \\
			\multicolumn{1}{c|}{}                     & MPANet~\cite{MPANet}       & 70.58 & \multicolumn{1}{c|}{68.24} & 75.58 & \multicolumn{1}{c|}{62.91} & 76.74 & \multicolumn{1}{c|}{80.95} & 84.22          & 75.11          & 82.8                    & \multicolumn{1}{c|}{80.7}    & 83.7                              & 80.9              \\ \hline
			\multicolumn{1}{c|}{\multirow{6}{*}{(2)}} & cmGAN~\cite{cmGAN}        & 26.97 & \multicolumn{1}{c|}{27.80} & 31.49 & \multicolumn{1}{c|}{22.27} & 31.63 & \multicolumn{1}{c|}{42.19} & 37.00          & 32.76          & -                       & \multicolumn{1}{c|}{-}       & -                                 & -                 \\
			\multicolumn{1}{c|}{}                     & AlignGAN~\cite{AlignGAN}     & 42.40 & \multicolumn{1}{c|}{40.70} & 51.50 & \multicolumn{1}{c|}{33.90} & 45.90 & \multicolumn{1}{c|}{54.30} & 57.10          & 45.30          & 65.36                   & \multicolumn{1}{c|}{53.40}   & 57.90                             & 53.60             \\
			\multicolumn{1}{c|}{}                     & cmPIG~\cite{cmPIG}        & 38.1  & \multicolumn{1}{c|}{36.9}  & 45.1  & \multicolumn{1}{c|}{29.5}  & 43.8  & \multicolumn{1}{c|}{52.9}  & 52.7           & 42.7           & 48.1                    & \multicolumn{1}{c|}{48.9}    & 48.5                              & 49.3              \\
			\multicolumn{1}{c|}{}                     & DDRL~\cite{DDRL}         & 29.80 & \multicolumn{1}{c|}{29.20} & -     & \multicolumn{1}{c|}{-}     & -     & \multicolumn{1}{c|}{-}     & -              & -              & 43.4                    & \multicolumn{1}{c|}{44.1}    & -                                 & -                 \\
			\multicolumn{1}{c|}{}                     & cm-SSFT~\cite{cm-SSFT}      & 47.70 & \multicolumn{1}{c|}{54.10} & -     & \multicolumn{1}{c|}{-}     & 57.40 & \multicolumn{1}{c|}{59.10} & -              & -              & 72.3                    & \multicolumn{1}{c|}{72.9}    & 71.0                              & 71.7              \\
			\multicolumn{1}{c|}{}                     & FMCNet~\cite{FMCNet}       & 66.34 & \multicolumn{1}{c|}{62.51} & 73.44 & \multicolumn{1}{c|}{56.06} & 68.15 & \multicolumn{1}{c|}{74.09} & 78.86          & 63.82          & \textcolor{red}{89.12}                   & \multicolumn{1}{c|}{\textcolor{red}{84.43}}   & \textcolor{red}{88.38}                             & 83.36             \\ \hline
			\multicolumn{2}{c|}{Flow2Flow-TSE}                       & 72.40 & \multicolumn{1}{c|}{69.77} & 77.26 & \multicolumn{1}{c|}{65.06} & 77.02 & \multicolumn{1}{c|}{81.24} & 84.23          & 76.60          & 85.63                   & \multicolumn{1}{c|}{83.14}   & 87.33                             & 84.24             \\
			\multicolumn{2}{c|}{Flow2Flow-CMG}                       & 71.75 & \multicolumn{1}{c|}{69.30} & 77.35 & \multicolumn{1}{c|}{65.08} & 77.66 & \multicolumn{1}{c|}{81.52} & 84.54          & 76.91          & 84.56                   & \multicolumn{1}{c|}{81.89}   & 86.12                             & 82.28             \\
			\multicolumn{2}{c|}{Flow2Flow}                           & \textcolor{red}{72.82} & \multicolumn{1}{c|}{\textcolor{red}{70.09}} & \textcolor{red}{77.75} & \multicolumn{1}{c|}{\textcolor{red}{65.84}} & \textcolor{red}{78.57} & \multicolumn{1}{c|}{\textcolor{red}{82.42}} & \textcolor{red}{84.58}          & \textcolor{red}{77.07}          & 86.02                   & \multicolumn{1}{c|}{83.08}   & 87.33                             & \textcolor{red}{83.60}             \\ \hline
		\end{tabular}
		\label{table_comparison}
	\end{table*}

	\begin{figure*}[t]
		\centering
		\subfigure[Baseline]{\includegraphics[width=0.245\textwidth]{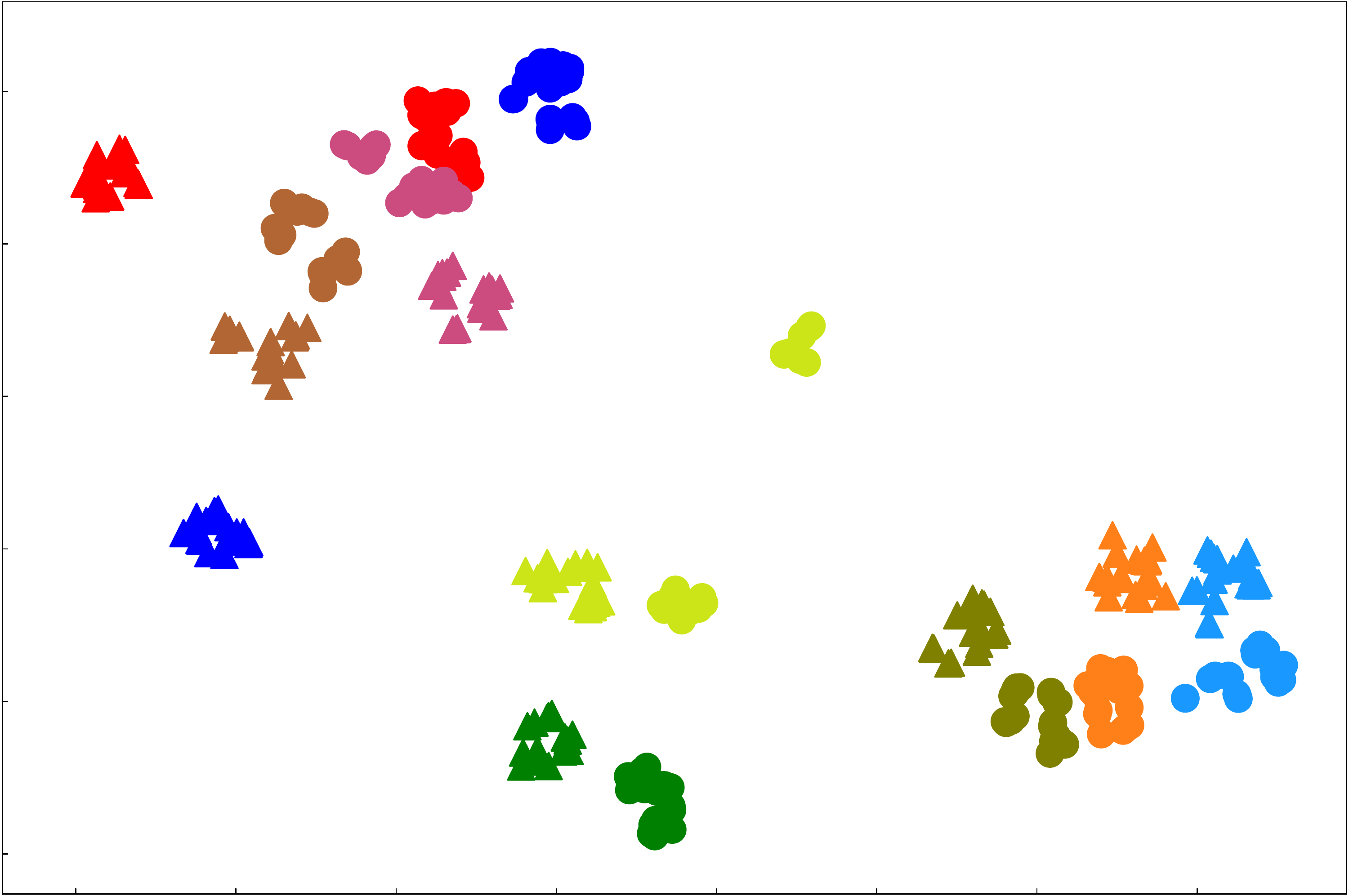}}
		\subfigure[Flow2Flow-TSE]{\includegraphics[width=0.245\textwidth]{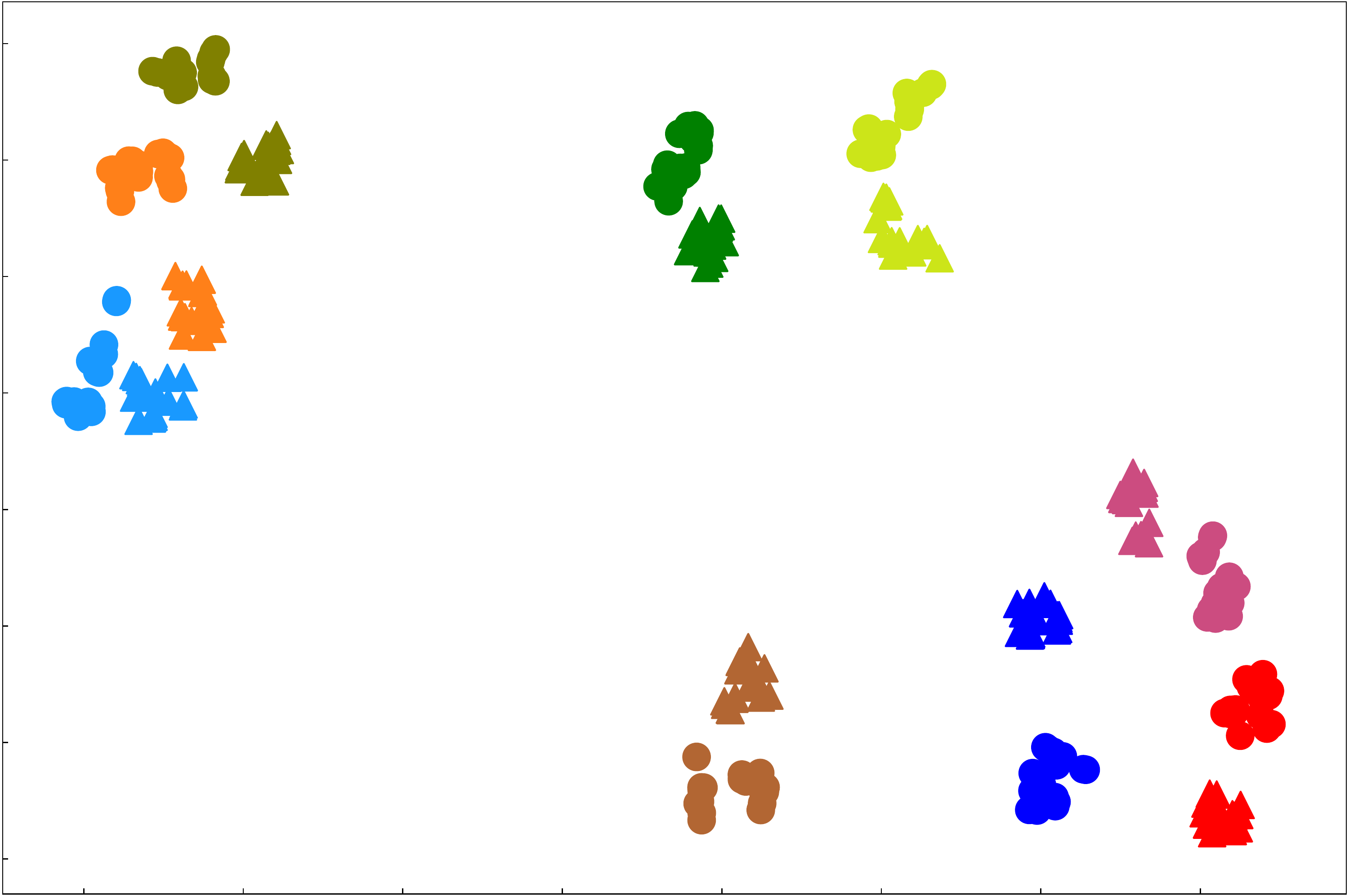}}
		\subfigure[Flow2Flow-CMG]{\includegraphics[width=0.245\textwidth]{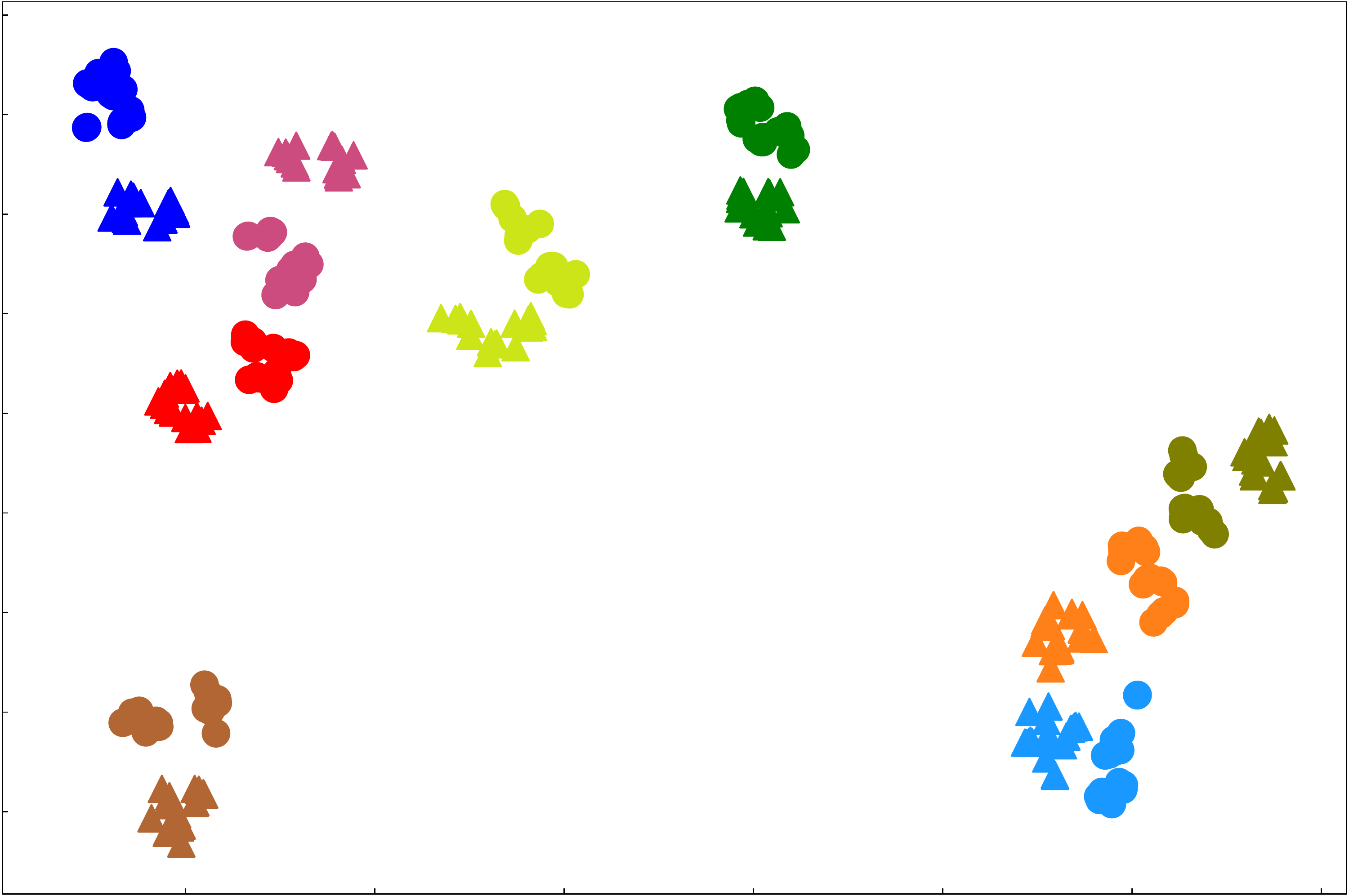}}
		\subfigure[Flow2Flow]{\includegraphics[width=0.245\textwidth]{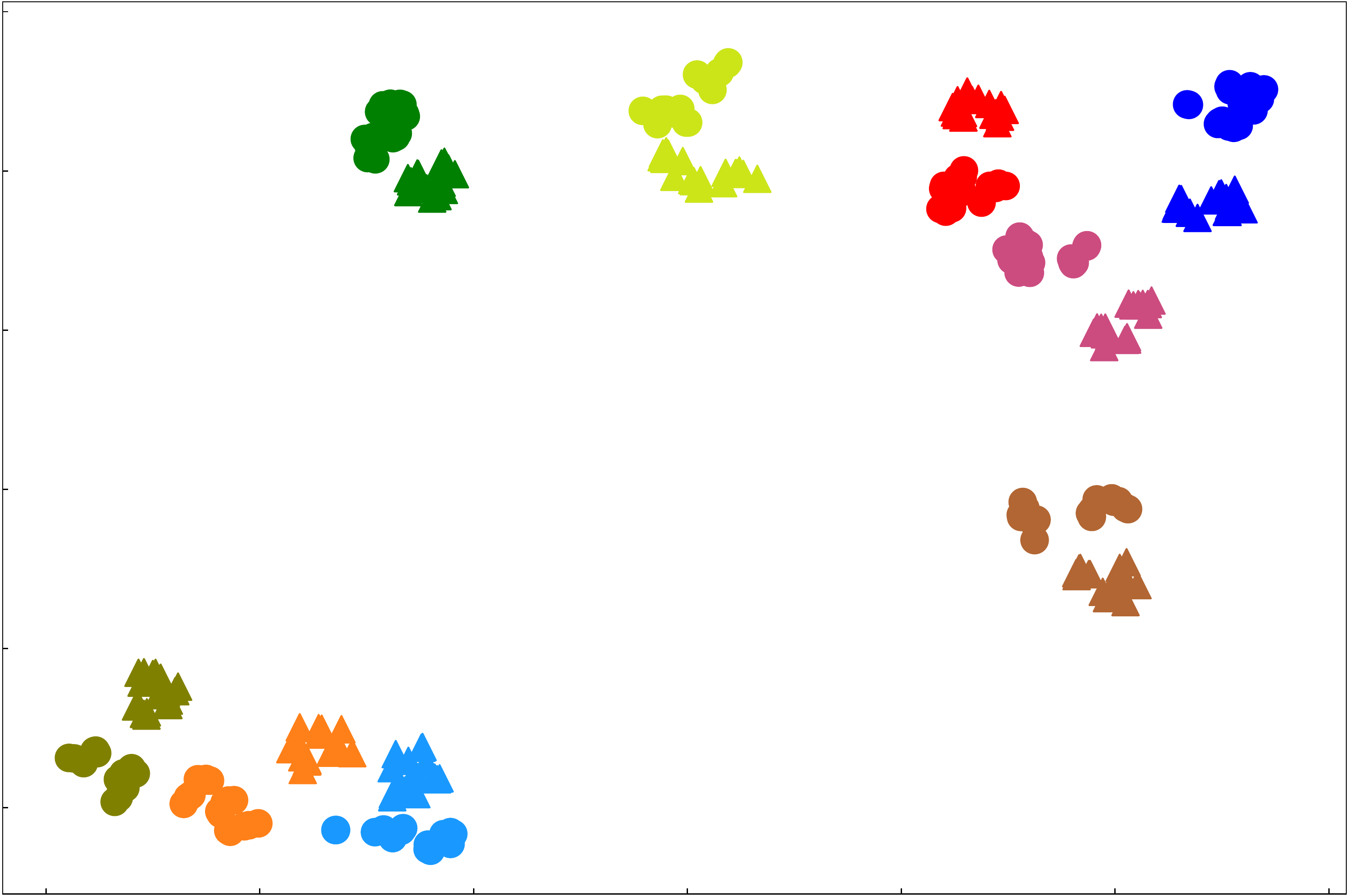}}
		\caption{Visualization of the pedestrian features.
			In this figure, marks with the same color belong to the same identity; the circle markers and triangle markers denote infrared features and visible features, respectively.
		}
		\label{fig_tsne}
	\end{figure*}

	\begin{figure*}[t]
	\centering
	\begin{center}
		\includegraphics[width=0.9 \textwidth]{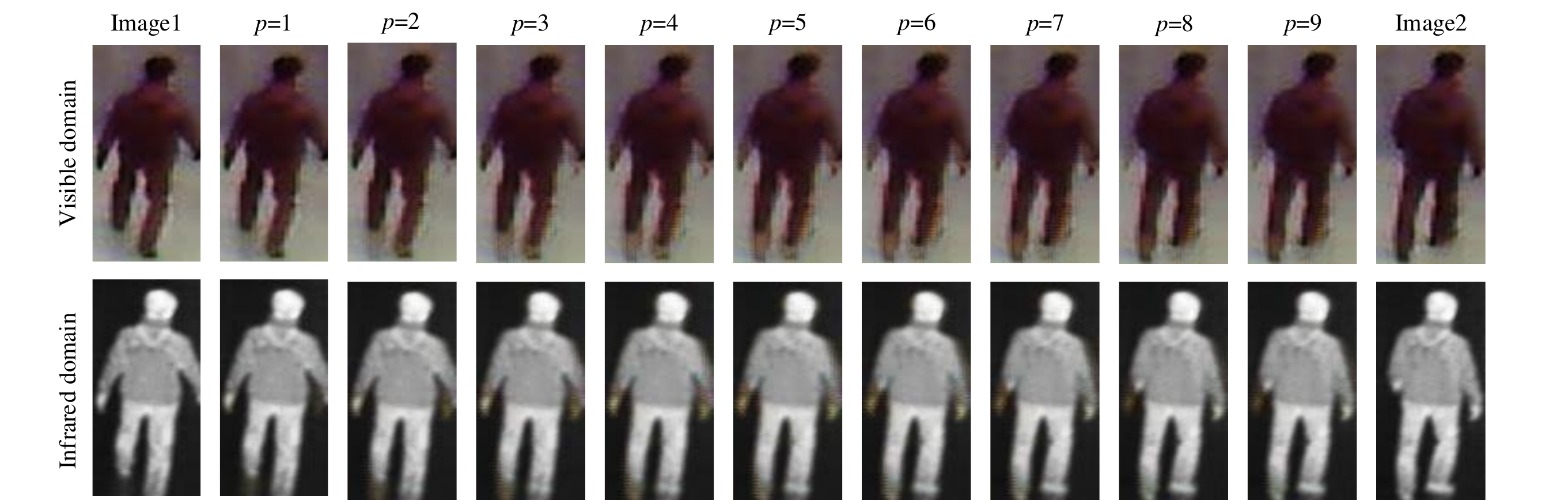}
	\end{center}
	\caption{Visualization of training sample generation by Flow2Flow-TSE.
		For the noise interpolation of Eq.~(\ref{eq_zv2xv_hat}), we set $q$ to 10 and $p \in \{1,2,3,4,5,6,7,8,9\}$.
	}
	\label{fig_vis_interpolation}
\end{figure*}

\begin{figure*}[t]
	\centering
	\begin{center}
		\includegraphics[width=0.9 \textwidth]{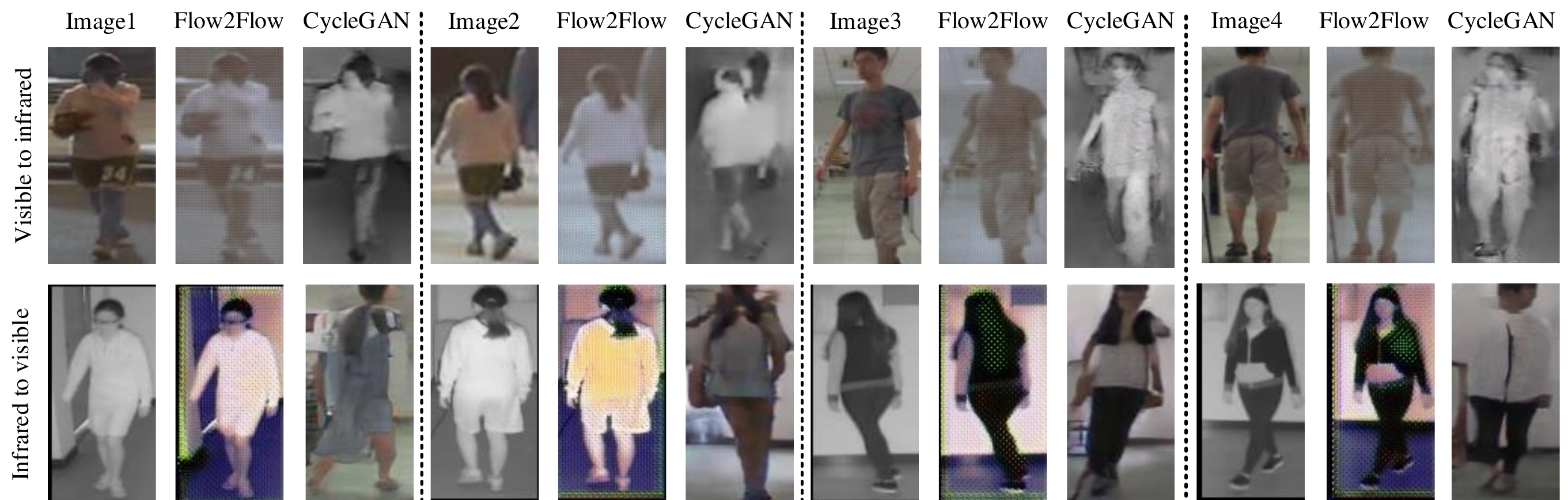}
	\end{center}
	\caption{Four groups of cross-modality image generation visualizations.
		In each group, the first column, second column and third column indicate the raw images, cross-modality images generated by Flow2Flow-CMG and cross-modality images generated by CycleGAN~\cite{CycleGAN}, respectively.
	}
	\label{fig_vis_CMG}
\end{figure*}

	\subsection{Comparison and Visualization}
	\label{comparison}
	
	\subsubsection{Quantitative comparison with SOTA}
	In this section, we compare our Flow2Flow to existing SOTA methods.
	Specifically, we compare two categories of methods: (1) modality-shared methods, such as expAT~\cite{expAT}, DDAG~\cite{DDAG} and MPANet~\cite{MPANet}; (2) modality compensation ones, such as cmGAN~\cite{cmGAN} and FMCNet~\cite{FMCNet}.
	We report the comparisons in Table~\ref{table_comparison}.
	In this table, \textbf{Flow2Flow-TSE} denotes the \textbf{T}raining \textbf{S}ample \textbf{E}xpansion; \textbf{Flow2Flow-CMG} represents the \textbf{C}ross-\textbf{M}odality \textbf{G}eneration; and \textbf{Flow2Flow} means that we perform both training sample expansion and cross-modality image generation.
	
	On SYSU-MM01~\cite{SYSU}, compared to the baseline model MPANet, the training sample expansion by Flow2Flow-TSE gains about 2.0\% improvement, while cross-modality image generation by Flow2Flow-CMG gains about 1.2\% improvement.
	Moreover, Flow2Flow with both training sample expansion and cross-modality image generation outperforms baseline by about 2.5\%, and thus leads a new state-of-the-arts (SOTA) performance.
	Hence, we could draw the following conclusions:
	1) the performance of V2I ReID is limited by the small number of training samples, and the training sample expansion by Flow2Flow could effectively alleviate this problem;
	2) the discrepancy between visible modality and infrared modality is one of the challenges of V2I ReID, while modality compensation by cross-modality image generation of Flow2Flow can reduce modality discrepancy;
	3) conducting training sample generation and cross-modality generation together is superior than single-type image generation.
	
	On RegDB, our Flow2Flow is only inferior than FMCNet~\cite{FMCNet} and thus achieves the second best performance.
	Compared to the baseline model, the training sample expansion by Flow2Flow-TSE could gain about 3.5\% improvement; while cross-modality image generation by Flow2Flow-CMG gains about 2.0\% improvement.
	Moreover, Flow2Flow with both training sample expansion and cross-modality image generation outperforms baseline by about 3.5\%, and achieve similar performance with Flow2Flow-TSE.
	For example, both Flow2Flow and Flow2Flow-TSE improves Rank1 of Visible2Infrared task from 83.7\% to 87.33\%.
	The training sample expansion on RegDB gains a higher improvement than that on SYSU-MM01.
	This is because the number of training samples in RegDB is much smaller than that of SYSU-MM01.

	\subsubsection{Qualitative analysis}
	In this section, we present visualizations for qualitative comparison.
	In Fig.~\ref{fig_tsne}, we first visualize the learned pedestrian features by baseline model, Flow2Flow-TSE, Flow2Flow-CMG and Flow2Flow.
	We employ t-SNE~\cite{tsne} to perform the data dimensionality reduction on pedestrian features.
	As can be seen, the features learned by baseline model suffer a huge intra-class modality discrepancy, especially for the red and blue markers.
	While cross-modality generation by Flow2Flow-CMG could effectively reduce the modality discrepancy.
	
	In Fig.~\ref{fig_vis_interpolation}, we visualize the pseudo training samples generated by latent noise interpolation.
	The value of $q$ in Eq.~(\ref{eq_zv2xv_hat}) is set to 10, $p$ is a positive integer from 1 to 9.
	In Fig.~\ref{fig_vis_CMG}, we visualize the cross-modality images generated by Flow2Flow-CMG and CycleGAN~\cite{CycleGAN}, a well-know algorithm for unsupervised pixel-level domain transformation.
	As can be seen, CycleGAN fails to model the pedestrian silhouette and appearance of given images.
	While Flow2Flow-CMG could achieve effective domain transformation.

	\subsection{Ablation Studies}
	\label{AS}
	In this section, we conduct multiple experiments on SYSU-MM01~\cite{SYSU} to verify each module proposed in this paper:
	we first compare our Flow2Flow-TSE to the baseline and training sample generation by image-sapce interpolation in Section~\ref{VoTSE};
	then compare Flow2Flow-CMG to baseline and CycleGAN-based cross-modality generation in Section~\ref{CMIG};
	next verify the impact of the nonlinear activation layer in Section~\ref{VoNAL};
	finally validate the effectiveness of the proposed adversarial training strategies and generator losses in Section~\ref{VoAT} and Section~\ref{VoGL}.

	\subsubsection{Verification of Training Sample Expansion}
	\label{VoTSE}
\begin{figure*}[t]
		\centering
		\subfigure[Visible image expansion]{\includegraphics[width=0.32\textwidth]{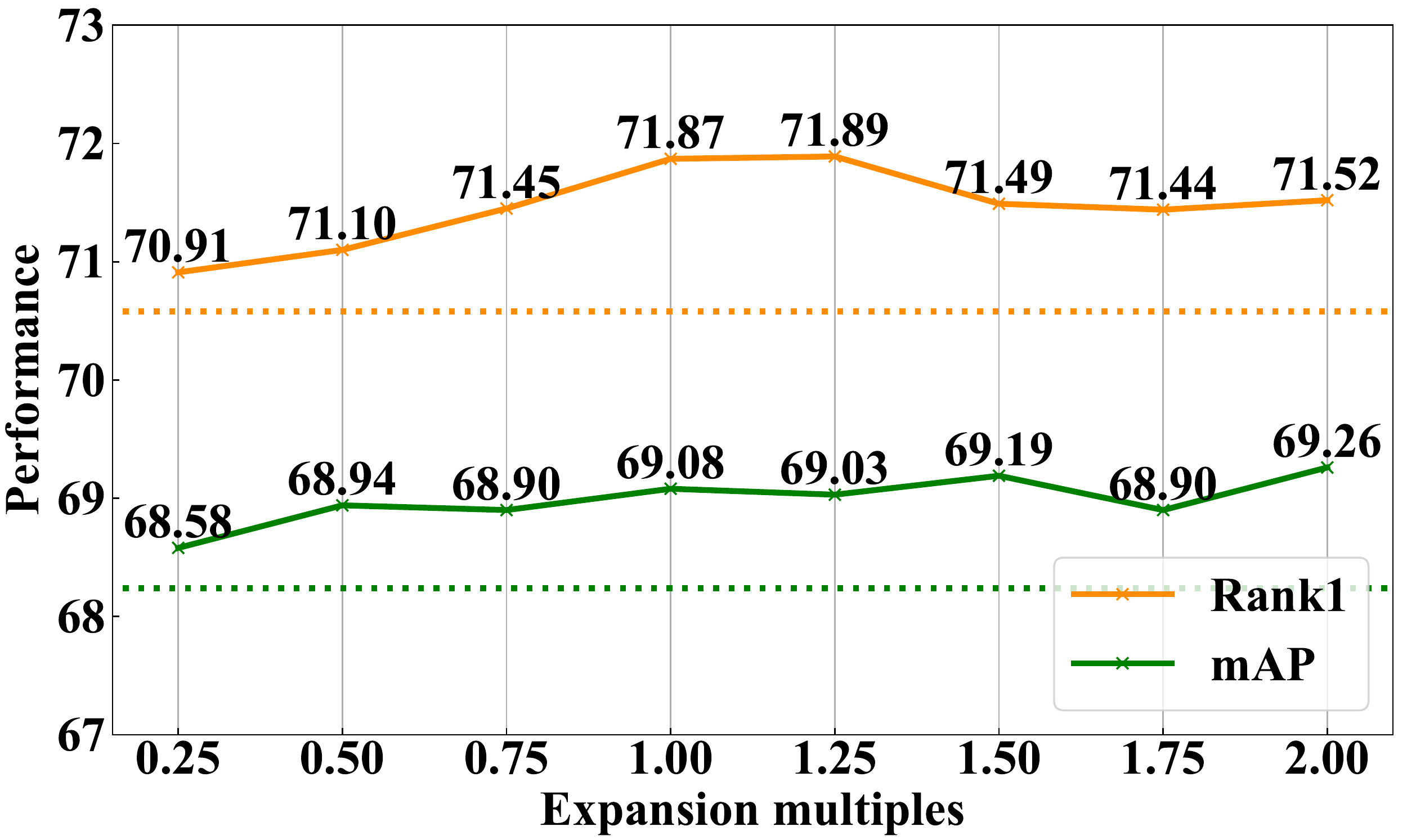}}
		\subfigure[Infrared image expansion]{\includegraphics[width=0.32\textwidth]{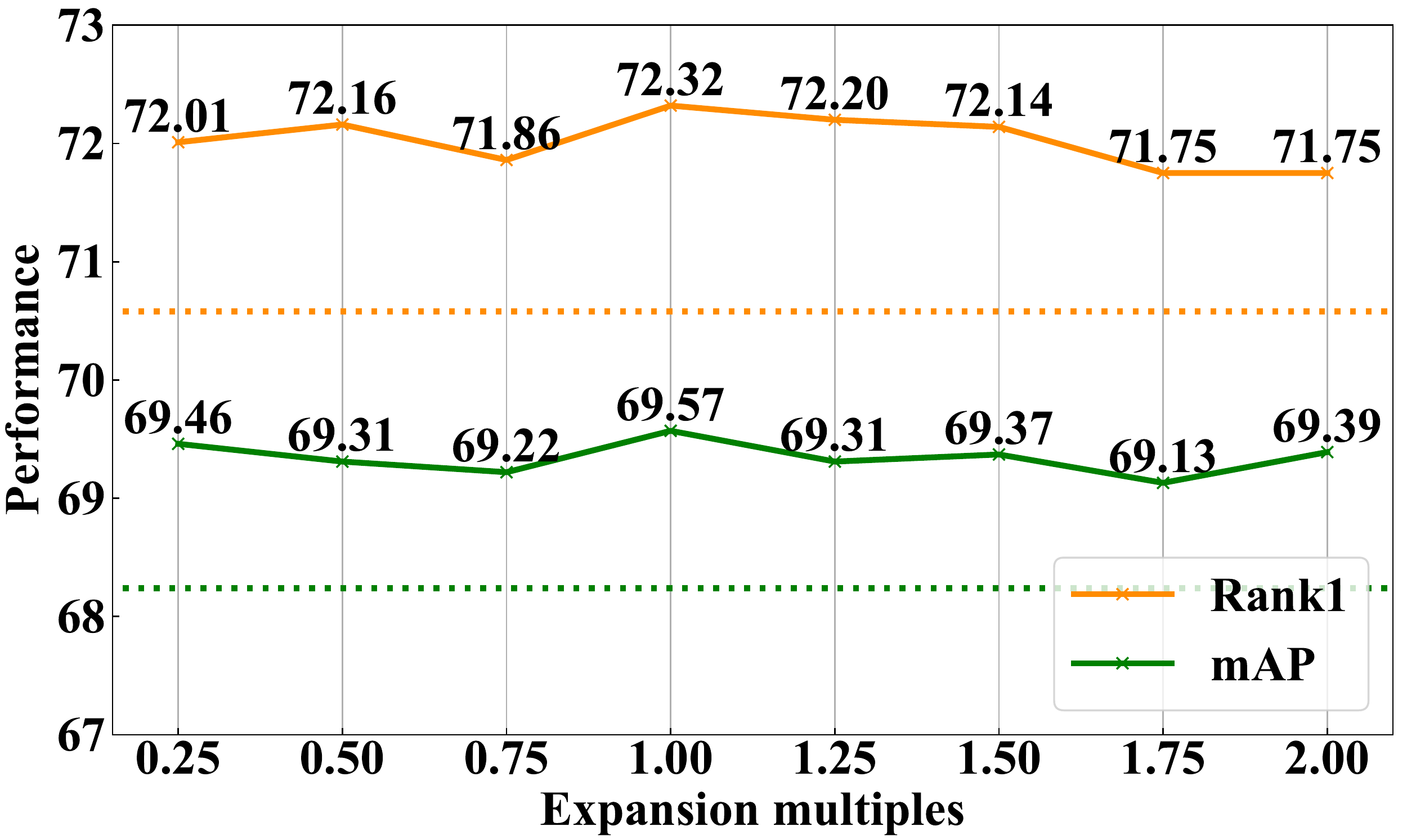}}
		\subfigure[Visible and infrared image expansion]{\includegraphics[width=0.32\textwidth]{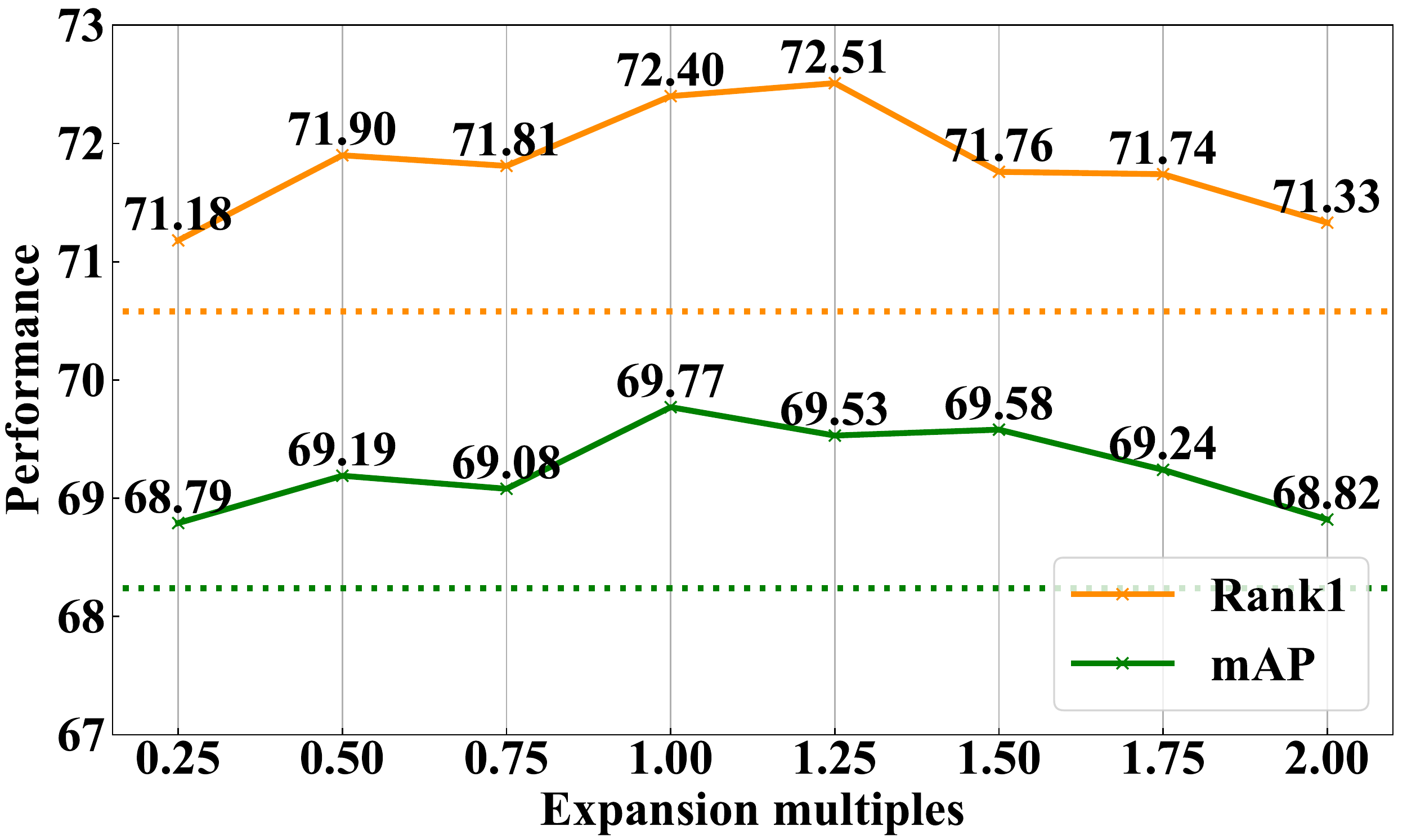}}
		\caption{Performance of training sample expansion, in which the dotted line denoted the performance of baseline model.
			X-axis represents the expansion multiple, for example, 1.5 means that the number of generated pseudo samples is 1.5 times that of true samples.
			In (a) or (b), we only generate the visible or infrared training images, while in (c), we generate both visible and infrared images.
		}
		\label{fig_expansion}
	\end{figure*}

	In this section, we verify the performance of Flow2Flow-TSE on SYSU-MM01~\cite{SYSU}.
	We could expand the training samples to multiples by Eq.~(\ref{eq_xv2zv}) and Eq.~(\ref{eq_zv2xv_hat}).
	We set $p$ and $q$ in Eq.~(\ref{eq_zv2xv_hat}) to 1 and 2, respectively.
	We test eight multiples for the generated samples, i.e., \{0.25, 0.5, 0.75, 1.0, 1.25, 1.5, 1.75, 2.0\}, and test the settings of 1) only expanding visible samples, 2) only expanding infrared images and 3) expanding both visible and infrared samples.
	The results are presented in Fig.~\ref{fig_expansion}.
	As can be seen, expansion on two modalities could achieve better performance than expansion on single modality, while accuracy of expanding infrared images is a slightly higher than that of expanding visible images, since SYSU-MM01 contains fewer infrared training samples.
	Meanwhile, in these three experiments, the best performance is obtained when the expansion multiple is equal to 1.
	This is because the proportion and efficacy of true training samples would decrease when the number of pseudo samples is too large.
	We conclude that Flow2Flow-TSE could gain a significant improvement compared to baseline model: Rank1 is improved from 70.58 to 72.40, while mAP is improved from 68.24 to 69.77.
	
	\begin{table}[t]
		\small
		\centering
		\renewcommand\arraystretch{1.1}
		\caption{Comparison with image-space linear interpolation and verification of $p$ and $q$.}
		\begin{tabular}{c|cc|cc}
			\hline
			\multirow{2}{*}{Method} & \multicolumn{2}{c|}{All-Search} & \multicolumn{2}{c}{Indoor-Search} \\ \cline{2-5} 
			& Rank1           & mAP           & Rank1            & mAP            \\ \hline
			Baseline                & 70.58              & 68.24                & 76.74               & 80.95                 \\
			Image Interpolation                    & 69.42              & 66.62                & 72.61               & 78.00                 \\
			Flow2Flow-TSE ($q=10$)             & 71.83              & 69.31                & \textcolor{red}{77.45}               & \textcolor{red}{81.47}                 \\
			Flow2Flow-TSE ($q=2$)              & \textcolor{red}{72.40}              & \textcolor{red}{69.77}                & 77.02               & 81.24                 \\ \hline
		\end{tabular}
		\label{table_TSE}
	\end{table}

\begin{figure}[t]
		\centering
		\begin{center}
			\includegraphics[width=0.44 \textwidth]{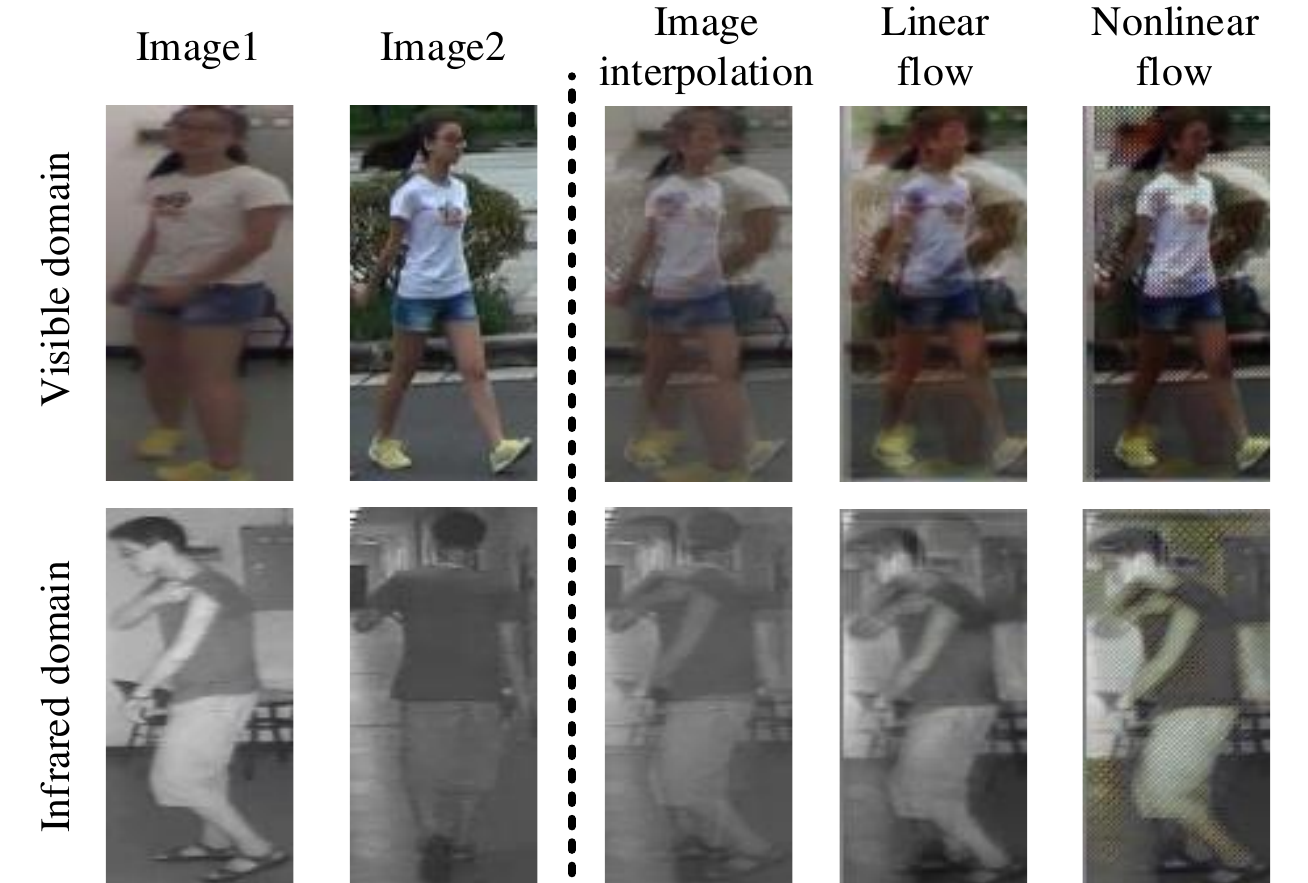}
		\end{center}
		\caption{Visualization of generated training samples.
			Images of column 1 and column 2 are true images;
			images of column 3 are generated by the image-space interpolation.
			images of column 3 and 4 are generated by latent noise interpolation ($p=1$ and $q=2$) of nonlinear flow and linear flow, respectively.
		}
		\label{fig_interpolation_linear}
	\end{figure}

	Furthermore, we compare Flow2Flow-TSE with the expansion by image-space linear interpolation, and verify other values for $p$ and $q$: $q=10$ while $p$ is randomly selected from \{1,2,3,4,5,6,7,8,9\}.
	We report the results in Table~\ref{table_TSE}.
	We find that training example expansion by image interpolation leads to worse performance, because it could not generate new valid samples, but reduces the proportion of true training samples.
	While for the latent noise interpolation, $q=10$ and $q=2$ obtain similar performance and they both outperform the baseline model.
	In Fig.~\ref{fig_interpolation_linear}, we visualize the generated training samples by Flow2Flow-TSE and image-space linear interpolation. Compared to the image interpolation, the latent noise interpolation can generate pseudo samples that are similar to true pedestrian images. For instance, in visible domain of Fig.~\ref{fig_interpolation_linear}, the fake image generated by noise interpolation of nonlinear flow retains the body shape of image 2 and dark style of image 1.

	\begin{table}[t]
		\small
		\centering
		\renewcommand\arraystretch{1.1}
		\caption{Comparison with baseline and cross-modality image generation by CycleGAN~\cite{CycleGAN}.}
		\begin{tabular}{c|cc|cc}
			\hline
			\multirow{2}{*}{Method} & \multicolumn{2}{c|}{All-Search} & \multicolumn{2}{c}{Indoor-Search} \\ \cline{2-5} 
			& Rank1           & mAP           & Rank1            & mAP            \\ \hline
			Baseline                & 70.58              & 68.24                & 76.74               & 80.95                 \\
			CycleGAN                    & 59.61              & 58.99                & 65.62               & 72.63                 \\
			Flow2Flow-CMG             & 71.75              & 69.30                & 77.66               & 81.52                 \\
			Flow2Flow              & \textcolor{red}{72.82}              & \textcolor{red}{70.09}                & \textcolor{red}{78.57}               & \textcolor{red}{82.42}                 \\ \hline
		\end{tabular}
		\label{table_CMG}
	\end{table}

	\subsubsection{Verification of Cross-Modality Image Generation}
	\label{CMIG}
	In this section, we verify the performance of Flow2Flow-CMG on SYSU-MM01~\cite{SYSU}.
	For each visible image $x_i^{<v>}$ in dataset, we generate a corresponding infrared one $\hat{x}^{<r>}_i$ by Eq.~(\ref{eq_xv2zv2xr}).
	And we generate a visible image $\hat{x}^{<v>}_i$ for $x_i^{<r>}$ by Eq.~(\ref{eq_xr2zv2xv}).
	We compare our Flow2Flow-CMG with CycleGAN-based cross-modality generation
	Their performances are reported in Table~\ref{table_CMG}.
	As can be seen, cross-modality generation by CycleGAN would greatly impair the experimental accuracy: both mAP and Rank1 are decreased by about 10\% compared to the baseline model.
	While Flow2Flow-CMG could gain about 1.2\% improvement on mAP and Rank1, which demonstrates the modality discrepancy would be reduced by cross-modality image generation of Flow2Flow.
	Moreover, Flow2Flow with both training sample expansion and cross-modality generation outperforms Flow2Flow.

	\begin{table}[t]
		\small
		\centering
		\renewcommand\arraystretch{1.1}
		\caption{Comparison of the linear flow and nonlinear flow for visible and infrared generators.}
		\begin{tabular}{c|c|cc|cc}
			\hline
			\multirow{2}{*}{Method}                                                   & \multirow{2}{*}{Linearity} & \multicolumn{2}{c|}{All-Search} & \multicolumn{2}{c}{Indoor-Search} \\ \cline{3-6} 
			&                            & mAP           & Rank1           & mAP            & Rank1            \\ \hline
			Baseline                                                                  & -                        & 70.58              & 68.24                & 76.74               & 80.95                 \\ \hline
			\multirow{2}{*}{\begin{tabular}[c]{@{}c@{}}Flow2Flow\\ -TSE\end{tabular}} & Linear                     &  70.80             & 67.96                & 75.53               & 80.12                 \\
			& Nonlinear                  & 72.40              & 69.77                & 77.02               & 81.24                 \\ \hline
			\multirow{2}{*}{\begin{tabular}[c]{@{}c@{}}Flow2Flow\\ -CMG\end{tabular}} & Linear                     & 71.35              & 69.16                & 77.69               & 80.93                 \\
			& Nonlinear                  & 71.75              & 69.30                & 77.66               & 81.52                 \\ \hline
			\multirow{2}{*}{Flow2Flow}                                                & Linear                     & 71.48              & 69.30                & 77.54               & 80.56                 \\
			& Nonlinear                  & \textcolor{red}{72.82}              & \textcolor{red}{70.09}                & \textcolor{red}{78.57}               & \textcolor{red}{82.42}                 \\ \hline
		\end{tabular}
		\label{table_linear}
	\end{table}

\begin{table}[t]
	\small
	\centering
	\renewcommand\arraystretch{1.1}
	\caption{Validation of the adversarial training strategies.
		In this table, ``none'' denotes that we omit the adversarial training; 
		``+ identity'' and ``+ modality'' denote that we only employ identity adversarial training and modality adversarial training to train Flow2Flow, respectively;
		and ``both'' means that we employ both identity adversarial training and modality adversarial training strategies.
	}
	\begin{tabular}{c|c|cc|cc}
		\hline
		\multirow{2}{*}{Method}                                                   & \multirow{2}{*}{\begin{tabular}[c]{@{}c@{}}Adversarial\\ Training\end{tabular}} & \multicolumn{2}{c|}{All-Search} & \multicolumn{2}{c}{Indoor-Search} \\ \cline{3-6} 
		&                                                                                 & mAP           & Rank1           & mAP            & Rank1            \\ \hline
		Baseline                                                                  & -                                                                               & 70.58              & 68.24                & 76.74               & 80.95                 \\ \hline
		\multirow{4}{*}{\begin{tabular}[c]{@{}c@{}}Flow2Flow\\ -TSE\end{tabular}} & none                                                                            & 72.19              & 69.32                & 76.90               & 81.09             \\
		& + identity                                                                       & 71.64              & 69.19                & 76.30               & 80.62                 \\
		& + modality                                                                       & 72.06              & 69.19                & 76.41               & 80.75                 \\
		& + both                                                              & 72.40              & 69.77                & 77.02               & 81.24                 \\ \hline
		\multirow{4}{*}{\begin{tabular}[c]{@{}c@{}}Flow2Flow\\ -CMG\end{tabular}} & none                                                                            & 67.78              & 64.04                & 73.31               & 77.43                 \\
		& + identity                                                                       & 70.23              & 67.11                & 75.73               & 79.92                 \\
		& + modality                                                                       & 69.65              & 66.83                & 75.50               & 79.74                 \\
		& + both                                                              & 71.75              & 69.30                & 77.66               & 81.52                 \\ \hline
		\multirow{4}{*}{Flow2Flow}                                                & none                                                                            & 72.39              & 69.41                & 77.45               & 81.38                 \\
		& + identity                                                                       & 72.29              & 69.25                & 76.96               & 81.05                 \\
		& + modality                                                                       & 72.00              & 69.21                & 77.08               & 81.14                 \\
		& + both                                                              & \textcolor{red}{72.82}              & \textcolor{red}{70.09}                & \textcolor{red}{78.57}               & \textcolor{red}{82.42}                 \\ \hline
	\end{tabular}
	\label{table_adversarial}
\end{table}

	\begin{figure*}[t]
		\centering
		\subfigure[Identity losses]{\includegraphics[width=0.3\textwidth]{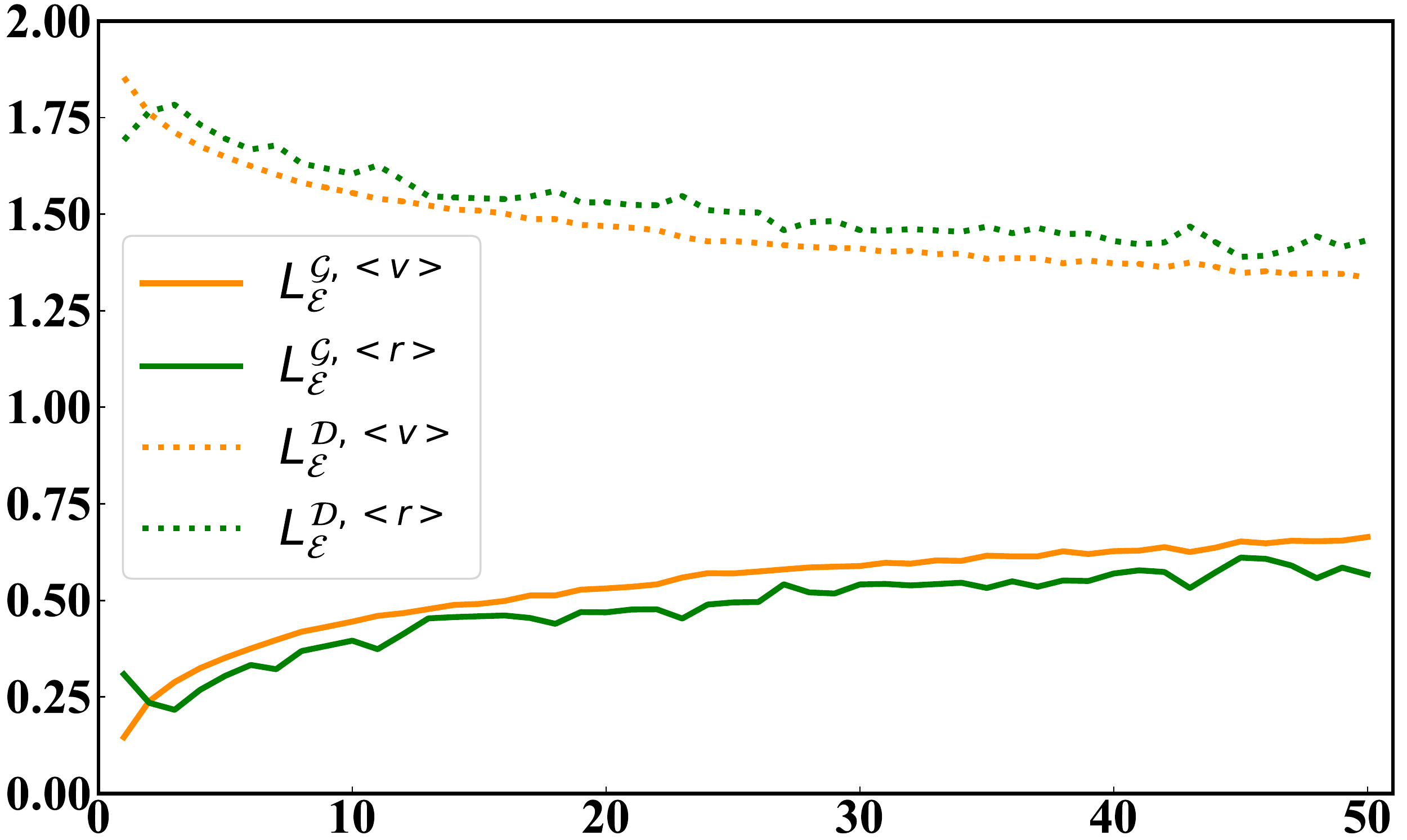} \label{fig_losses_a}}
		\subfigure[Modality losses]{\includegraphics[width=0.3\textwidth]{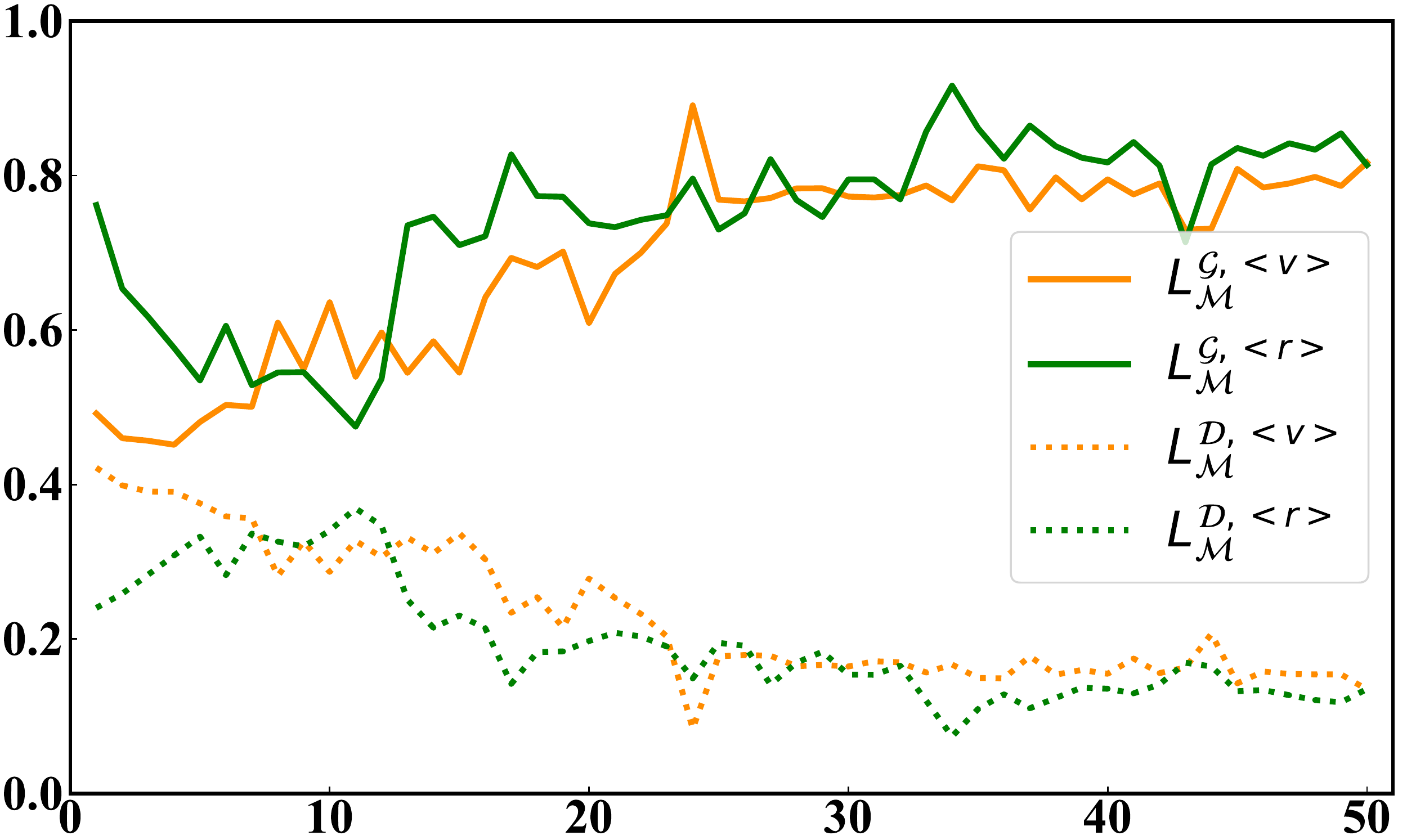}\label{fig_losses_b}}
		\subfigure[Flow losses and latent loss]{\includegraphics[width=0.3\textwidth]{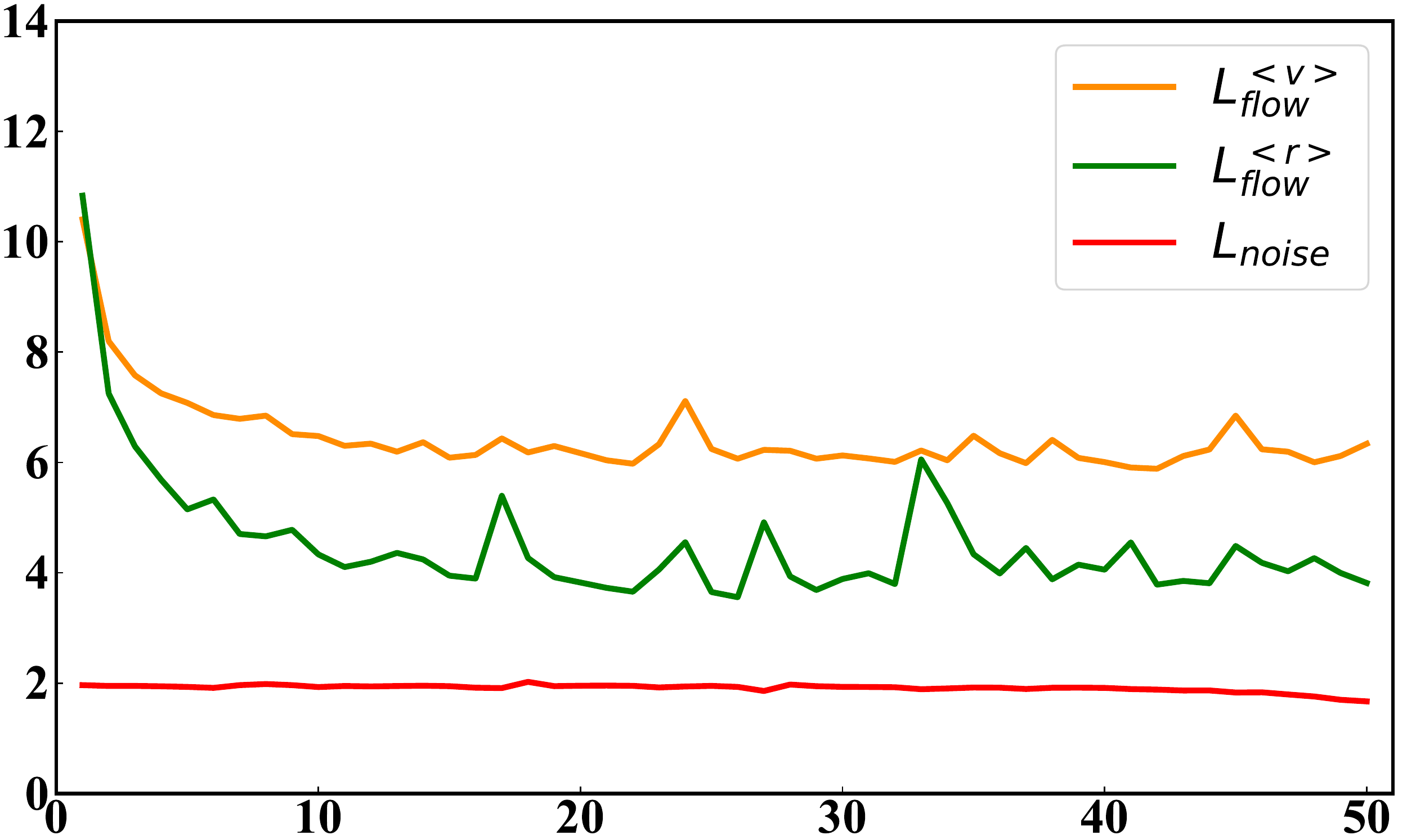}\label{fig_losses_c}}
		\caption{The visualization of training losses.
			In this figure, the orange lines and green lines denote losses of visible domain and infrared domain, respectively.
			In (a) and (b), the solid line and dotted line indicate losses of generator and discriminator, respectively.
		}
		\label{fig_losses}
	\end{figure*}

\begin{figure}[ht]
	\centering
	\begin{center}
		\includegraphics[width=0.44 \textwidth]{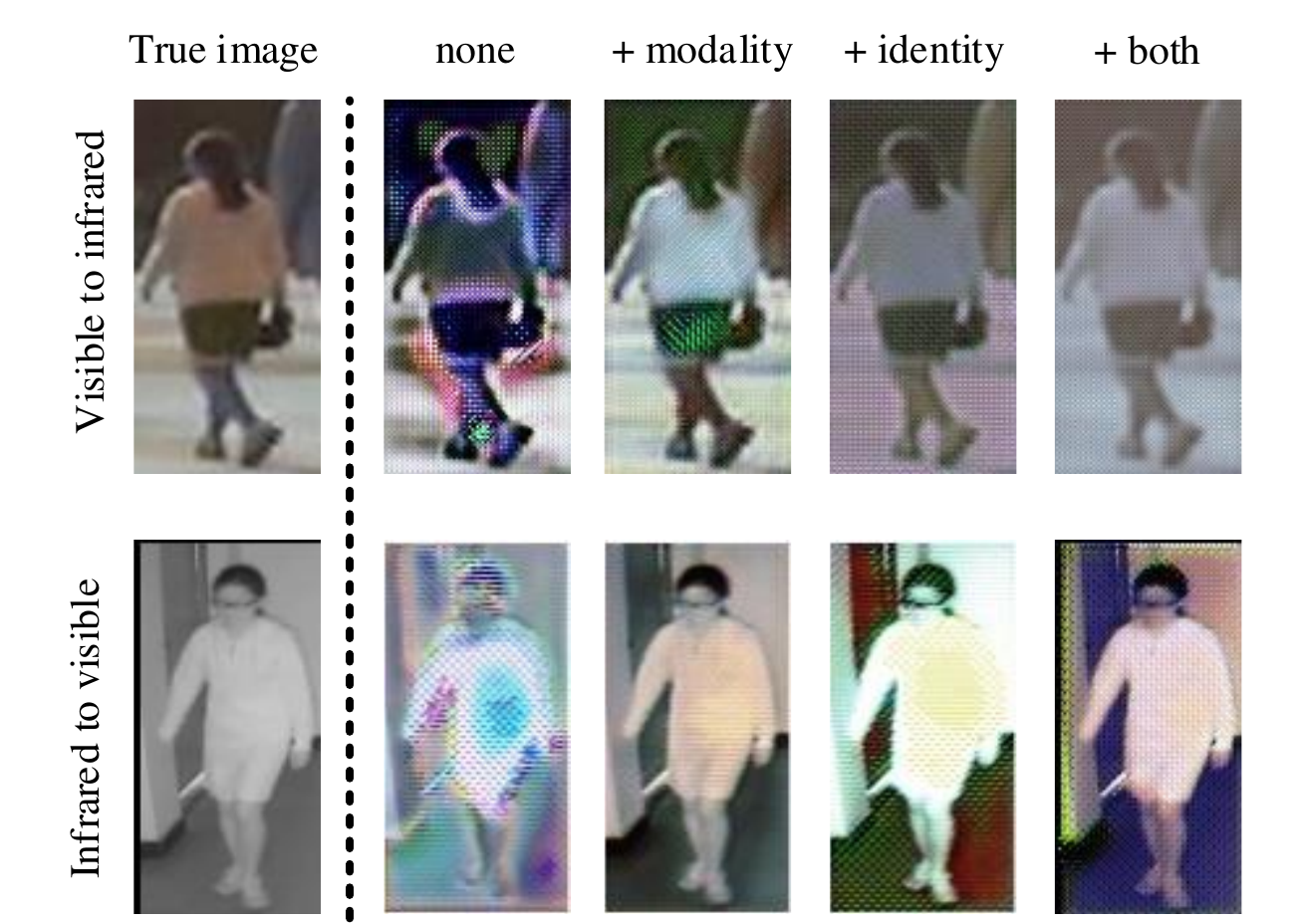}
	\end{center}
	\caption{Visualization of cross-modality generation, in which images of column 1 are true images.
		In this figure, ``none'' denotes that we omit the adversarial training;
		``+ modality'' and ``+ identity'' denote that we only employ modality adversarial training and identity adversarial training to train Flow2Flow, respectively;
		and ``both'' means that we employ both identity adversarial training and modality adversarial training strategies;
	}
	\label{fig_adversarial_training}
\end{figure}

	\subsubsection{Verification of Nonlinear Activation Layer}
	\label{VoNAL}
	In this section, we verify the proposed nonlinear activation layer of Eq.~(\ref{eq_nonlinear_rev}) and Eq.~(\ref{eq_nonlinear_for}).
	We test Flow2Flow model with or without nonlinear layer and report results in Table~\ref{table_linear}.
	As can be seen, for the training sample expansion by Flow2Flow-TSE, the linear flow is inferior to the nonlinear flow and only achieves similar performance with baseline model.
	While for the Flow2Flow-CMG, the performance is slightly affected by the linearity of flow.
	The analysis is presented as follows.
	Flow2Flow-TSE generates pseudo training samples by linear interpolation of latent noises, thereby, it would fail to generate new effective samples if the nonlinearty of generators is insufficient.
	From the visualization in Fig.~\ref{fig_interpolation_linear}, we find that pseudo samples generated by linear flow are similar to the results of image interpolation.

	\begin{table}[ht]
		\small
		\centering
		\renewcommand\arraystretch{1.1}
		\caption{Validation of the generator loss.
			In this table, ``none'' denotes that we omit the flow loss and noise loss; 
			``+ flow'' and ``+ noise'' denote that we only employ flow loss and noise loss to train Flow2Flow, respectively;
			and ``+ both'' means that we employ both flow loss and noise loss to train model.
		}
		\begin{tabular}{c|c|cc|cc}
			\hline
			\multirow{2}{*}{Method}                                                   & \multirow{2}{*}{\begin{tabular}[c]{@{}c@{}}Generator\\ Losses\end{tabular}} & \multicolumn{2}{c|}{All-Search} & \multicolumn{2}{c}{Indoor-Search} \\ \cline{3-6} 
			&                                                                                 & mAP           & Rank1           & mAP            & Rank1            \\ \hline
			Baseline                                                                  & -                                                                               & 70.58              & 68.24                & 76.74               & 80.95                 \\ \hline
			\multirow{4}{*}{\begin{tabular}[c]{@{}c@{}}Flow2Flow\\ -TSE\end{tabular}} & none                                                                            & 69.85              & 67.04                & 74.97               & 79.67                 \\
			& + flow                                                                       & 71.43              & 68.96                & 76.38               & 80.78                 \\
			& + noise                                                                       & 70.65              & 67.91                & 76.44               & 80.52                 \\
			& + both                                                              & 72.40              & 69.77                & 77.02               & 81.24                 \\ \hline
			\multirow{4}{*}{\begin{tabular}[c]{@{}c@{}}Flow2Flow\\ -CMG\end{tabular}} & none                                                                                         & 69.28                & 67.42               & 74.85   & 79.36              \\
			& + flow                                                                       & 70.44              & 68.55                & 76.35               & 80.25                 \\
			& + noise                                                                       & 69.89              & 67.68                & 75.76               & 79.63                 \\
			& + both                                                              & 71.75              & 69.30                & 77.66               & 81.52                 \\ \hline
			\multirow{4}{*}{Flow2Flow}                                                & none                                                                            & 70.17               & 68.28                & 75.42               & 79.52                 \\
			& + flow                                                                       & 72.02              & 69.31                & 76.99               & 80.86                 \\
			& + noise                                                                       & 70.19              & 68.17                & 76.16               & 80.15                 \\
			& + both                                                              & \textcolor{red}{72.82}              & \textcolor{red}{70.09}                & \textcolor{red}{78.57}               & \textcolor{red}{82.42}                 \\ \hline
		\end{tabular}
		\label{table_generator_loss}
	\end{table}

	\subsubsection{Verification of Adversarial Training}
	\label{VoAT}
	In this section, we verify the impact of adversarial training strategies on image generation.
	We test multiple experimental settings on Flow2Flow: no adversarial training, adopting identity adversarial training only, adopting modality adversarial training only, and adopting both identity and modality adversarial training.
	We present the results in Table~\ref{table_adversarial}.
	For the training sample expansion by Flow2Flow-TSE, the adversarial training slightly affects the ReID accuracy: the performance of Flow2Flow-TSE with no adversarial training strategy is similar to that of Flow2Flow-TSE with two adversarial training strategies.
	Thanks to the invertibility of flow-based generative models, the interpolated latent noises can always generate valid training samples with or without adversarial training.
	While for cross-modality generation by Flow2Flow-CMG, the adversarial training strategies plays an important role: Flow2Flow-CMG with two adversarial training strategies outperforms Flow2Flow-CMG with no adversarial training strategy by about 4\%.
	Moreover, adopting both identity and modality adversarial training is superior than single adversarial training.
	
	In Fig.~\ref{fig_adversarial_training}, we visualize the cross-modality images generated by Flow2Flow-CMG with and without adversarial training.
	As can be seen, Flow2Flow-CMG with both identity and modality adversarial training could generate high-quality cross-modality images, while Flow2Flow-CMG with no adversarial training fails to capture the modality information and pedestrian appearance.
	In Fig.\ref{fig_losses_a} and Fig.\ref{fig_losses_b}, we visualize the identity losses and modality losses druing adversarial training, respectively.
	We conclude that the generator losses and discriminator losses reach the adversarial equilibrium.

	\subsubsection{Verification of Generator Losses}
	\label{VoGL}
	In this section, we verify the impact of generator losses $L_{flow}$ and $L_{noise}$.
	We test multiple experimental settings on Flow2Flow: model training without $L_{flow}$ and $L_{noise}$, with only $L_{flow}$, with only $L_{noise}$, and with both generator losses.
	From the results of Table~\ref{table_generator_loss}, we find that both flow loss and noise loss have a positive impact on performance of Flow2Flow, and the flow loss is more important than the noise loss.
	In Fig.\ref{fig_losses_c} we visualize $L_{flow}$ and $L_{noise}$ during training Flow2Flow.
	Compared to the visible flow loss, the infrared flow loss could converge to a lower value.
	The single-channel infrared images contains less color and texture information, therefore it would be easier to transform the infrared images to Gaussian noises.

	\section{Conclusions}
	\label{conclusion}
	In this paper, we verified how image generation, including training sample generation and cross-modality generation, helps the visible-to-infrared person ReID.
	To this end, we proposed a unified framework, named Flow2Flow, to jointly achieve training sample expansion and cross-modality image generation.
	Flow2Flow consists of a visible flow and an infrared flow, which transform the visible images and infrared images to isotropic noises, respectively.
	Thus, we could generate new training samples and cross-modality images by using the invertibility of flow-based models.
	Moreover, an image encoder and a modality discriminator were devised for identity adversarial training and modality adversarial training, respectively.
	Experimental results on SYSU-MM01 and RegDB demonstrated that both training sample expansion and cross-modality image generation could improve the performance V2I ReID, in which training sample expansion could gain a higher improvement.
	Thereby, we could draw conclusion from this paper that both the lack of training samples and cross-modality discrepancy limited the accuracy of V2I person ReID.

	\bibliographystyle{IEEEtran}
	\bibliography{IEEEabrv,mybibfile}
	
\end{document}